\documentclass[11pt, table]{article}
\def\bng{\bngx}

\font\bngx=bang10

\def\*#1*#2{o\null{#2}{#1}}

\def\sh#1{\setbox0=\hbox{#1}%
     \kern-.02em\copy0\kern-\wd0
     \kern.04em\copy0\kern-\wd0
     \kern-.02em\raise.0433em\box0 }

\usepackage[final]{acl}

\usepackage{times}
\usepackage{latexsym}
\usepackage{multirow}
\usepackage{subcaption}
\usepackage{mathtools}
\usepackage{amsmath}
\usepackage[T1]{fontenc}
\usepackage[utf8]{inputenc}

\usepackage{microtype}

\usepackage{inconsolata}
\usepackage{graphicx}

\usepackage{multirow}
\usepackage{xcolor}
\usepackage{booktabs}
\usepackage{amsmath}
\usepackage{tabularray}
\usepackage{makecell}
\usepackage{enumitem}
\title{Mitigating Extrinsic Gender Bias for Bangla Classification Tasks}

\author{
 \textbf{Sajib Kumar Saha Joy\textsuperscript{*,1}},
 \textbf{Arman Hassan Mahy\textsuperscript{*,1}},
 \textbf{Meherin Sultana\textsuperscript{1}},
 \\
 \textbf{Azizah Mamun Abha\textsuperscript{1}},
 \textbf{MD Piyal Ahmmed\textsuperscript{1}},
 \textbf{Yue Dong\textsuperscript{2}},
 \textbf{G M Shahariar\textsuperscript{2}}
\\
 \textsuperscript{1} Ahsanullah University of Science and Technology\\
 \textsuperscript{2} University of California, Riverside
\\
\texttt{joy.cse@aust.edu}\\
\texttt{\{say2mahy, sultanameh01, abha.azizah800, yasinarafath21\}@gmail.com}\\
\texttt{\{yue.dong, gshah010\}@ucr.edu}\\
}

\begin{document}
\maketitle

\renewcommand{\thefootnote}{\fnsymbol{footnote}}
\footnotetext[1]{These authors contributed equally to this work.}
\renewcommand{\thefootnote}{\arabic{footnote}}

\begin{abstract}
In this study, we investigate extrinsic gender bias in Bangla pretrained language models, a largely underexplored area in low-resource languages. To assess this bias, we construct four manually annotated, task-specific benchmark datasets for sentiment analysis, toxicity detection, hate speech detection, and sarcasm detection. Each dataset is augmented using nuanced gender perturbations, where we systematically swap gendered names and terms while preserving semantic content, enabling minimal-pair evaluation of gender-driven prediction shifts. We then propose \textit{RandSymKL}, a randomized debiasing strategy integrated with symmetric KL divergence and cross-entropy loss to mitigate the bias across task-specific pretrained models. \textit{RandSymKL} is a refined training approach to integrate these elements in a unified way for extrinsic gender bias mitigation focused on classification tasks. Our approach was evaluated against existing bias mitigation methods, with results showing that our technique not only effectively reduces bias but also maintains competitive accuracy compared to other baseline approaches. To promote further research, we have made both our implementation and datasets publicly available.\footnote{\url{https://github.com/sajib-kumar/Mitigating-Bangla-Extrinsic-Gender-Bias}}
\end{abstract}

\section{Introduction}
Large Language Models (LLMs) such as GPT-4~\cite{achiam2023gpt}, Gemini~\cite{team2023gemini}, LLaMA~\cite{touvron2023llama} and their successors have revolutionized text generation and understanding by leveraging vast amounts of data and substantial computational power to achieve state-of-the-art performance. However, their sheer size and resource requirements make them impractical for many real-world applications in resource-constrained languages, such as Bangla, where the computational cost of deploying LLMs can be prohibitive \cite{hasan2024large}. As a result, task specific Pretrained Language Models (PLMs) such as BERT~\cite{devlin-etal-2019-bert} and T5~\cite{raffel2020exploring}, which are more lightweight and computationally efficient are often favored for their balance between performance and practicality, making them suitable for various NLP and NLG tasks in low resource language like Bangla \cite{kabir-etal-2024-benllm}. Despite their widespread adoption in classification tasks, one significant issue in utilizing PLMs is the presence of bias towards gender, which can lead to skewed and potentially harmful outcomes~\cite{NIPS2016_a486cd07}.
\begin{figure}
    \centering
    \vspace{0.25cm}
    \includegraphics[width=1\linewidth]{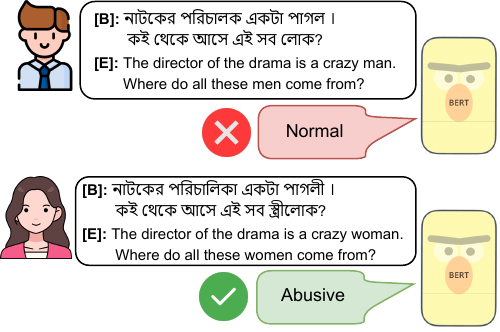}
    \caption{An example of prediction mismatch in hate speech classification task using a pretrained BERT-based model. The female-centric sentence is correctly classified as abusive, while its semantically equivalent male-centric counterpart is misclassified as normal.}
    \vspace{-17pt}
    \label{fig:example_bias}
\end{figure}

Numerous definitions of gender bias have been proposed in the NLP literature, reflecting different theoretical and methodological perspectives ~\cite{gallegos-etal-2024-bias}. However, our work focuses whether PLMs exhibit unequal predictive behavior across gendered text variants. Specifically, we examine whether PLMs predict male-centric sentences more accurately than their semantically equivalent female-centric counterparts, or vice versa. Such disparities arise from prediction mismatches between gender-swapped sentence pairs, where semantically equivalent inputs receive different model predictions, as shown in Figure~\ref{fig:example_bias}. Following the taxonomy of gender biases in NLP proposed by \citet{gallegos-etal-2024-bias}, this phenomenon constitutes a form of representational harm, specifically disparate system performance, where a model demonstrates unequal understanding or predictive accuracy across linguistic realizations associated with different social groups.

While such gender bias has been extensively studied in English across a range of NLP tasks ~\cite{gallegos-etal-2024-bias}, both in terms of detection~\cite{kiritchenko-mohammad-2018-examining} and mitigation~\cite{sun-etal-2019-mitigating}, research in Bangla remains limited. Despite the increasing availability of pretrained Bangla language models, existing studies~\cite{sadhu2024empirical, sadhu2024social} have primarily focused on intrinsic gender bias, that is, bias encoded within model representations, whereas extrinsic bias, reflected through unequal predictive behavior in downstream classification tasks, remains largely unexplored.

This paper addresses the gap of extrinsic bias mitigation by examining gender bias in four different Bangla language models pretrained in classification tasks of sentiment analysis, toxicity detection, hate speech detection, and sarcasm detection. We first curated four manually annotated task-specific datasets, containing male-centric and corresponding female-centric text, specifically transformed to identify extrinsic gender bias by swapping names and gender-specific terms, since previous work has shown that such manipulation of data can reveal latent biases in language models \cite{rudinger-etal-2018-gender}. Then, we applied several gender bias mitigation techniques such as data augmentation, counterfactual data substitution, fine-tuning with bias-neutral datasets, and embedding and probability distribution alignment of text pairs that have been explored in prior research \cite{zhao-etal-2018-gender, DBLP:journals/corr/abs-2010-06032} to assess their effectiveness in reducing bias while maintaining classification performance. 

In addition, we propose a refined debiasing training strategy, \textit{RandSymKL}, which jointly optimizes cross-entropy loss and symmetric KL divergence loss, and compared the performance with the existing methods. Our findings show that RandSymKL slightly outperforms existing methods across all bias evaluation metrics while preserving task-specific classification performance, demonstrating its potential application in broader gender bias free classification tasks. We summarize our main contributions below:
\begin{enumerate}[leftmargin=15pt, topsep=2pt, itemsep=0pt]
    \item We construct a comprehensive Bangla-specific lexicon of 573 gendered word pairs to support precise counterfactual data generation for bias evaluation. Using this lexicon, we then generate counterfactual gender-swapped text to systematically assess model behavior across gendered variations.
    
    \item \textbf We propose a stochastic debiasing method, RandSymKL, which combines randomized cross-entropy loss with symmetric KL divergence to align predictions across gender-swapped text pairs. Our experiments demonstrate that RandSymKL effectively balances bias mitigation and task performance, outperforming existing debiasing baselines.
    
    \item To the best of our knowledge, this study presents the first comprehensive investigation of extrinsic gender bias mitigation for four different Bangla text classification tasks. We also release curated gender-swapped datasets to facilitate benchmarking.
\end{enumerate}

\section{Prior Works}
Gender bias can manifest in both intrinsic and extrinsic forms. Various techniques have been proposed to measure and mitigate bias accurately.

\subsection{Intrinsic Gender Bias Detection Methods}
 \citet{sun-etal-2019-mitigating} examined intrinsic gender bias through the Implicit Association Test (IAT) and Word Embedding Association Test (WEAT) \citet{Caliskan_2017},  applied to GloVe and Word2Vec embeddings on the GBETs dataset. \citet{tiwari2022casteism} explored intrinsic gender bias in Hindi and Tamil monolingual word embeddings using the WEAT metric, which aligned with human bias, while also assessing caste bias. \citet{tokpo2023far} used the Sentence Embedding Association Test (SEAT) and Log Probability Bias Score (LPBS).

 \subsection{Extrinsic Gender Bias Detection Methods}
\citet{su2023learning} proposed a method to detect extrinsic gender bias in LLMs such as Alpaca, ChatGPT, and GPT-4, by generating gender-swapped sentence pairs and checking for contradictions in LLM responses. Bias is identified when outputs differ despite semantically equivalent inputs. \citet{jentzsch2023gender} revealed significant extrinsic gender biases in BERT models using the IMDB dataset by comparing sentiment classification results for gender-swapped versions of movie reviews and analyzing the sentiment score disparities between them.  In addition to intrinsic bias, \citet{tokpo2023far} also measured extrinsic bias using True Positive Rate Difference between the predictions for two groups, and Counterfactual Fairness between test examples and their counterfactual one. \citet{xie2023countergap} proposed Counter-GAP, a counterfactual framework to evaluate extrinsic gender bias in coreference resolution by constructing gender-swapped sentence quadruples. Bias was measured using Inconsistency Across Genders (IAG), Inconsistency Within Genders (IWG), and a derived bias score (IAG \text{--} IWG), revealing notable inconsistencies in BERT-based models.  

\citet{chen2024bias} examined both intrinsic and extrinsic gender bias through statistical and causal fairness frameworks, using metrics such as Statistical Parity Difference (SPD), Equal Opportunity Difference (EOD), and Counterfactual Fairness. They used statistical frameworks to compare prediction outcomes between groups (male and female) and a causal framework to compare counterfactual predictions. \citet{qian-etal-2022-perturbation} introduced a demographic bias detection method in NLP, using perturbation augmentation and FairScore to evaluate fairness based on prediction stability. \citet{sobhani-delany-2024-towards} introduced methods to detect gender bias by analyzing prediction disparities tied to identity terms across real-world datasets, using accuracy, F1 score differences to measure bias presence in classification models like hate speech detection and sentiment analysis.

% \citet{das2023toward} focused on gender and racial biases in Bangla NLP systems using the BIBED model and RDF, analyzing a diverse set of sentences from Wikipedia, Banglapedia, Bangla literature, and other sources.

\subsection{Bias Mitigation Methods}
\citet{hort2023bias} identified three types of bias mitigation methods: Pre-processing, In-processing, and Post-processing. These methods aim to ensure fairness in machine learning models (ML). \citet{tokpo2023far} proposed strategies to mitigate gender bias in contextualized language models such as BERT-large and ALBERT-large, focusing on text classification tasks. Techniques like CDA pretraining, Context-derbies, and Sent-derbies were explored, addressing bias in pretraining data, procedures, or model outputs. Similarly, \citet{su2023learning} addressed bias in Natural Language Generation (NLG) using algorithm-based Adversarial Learning and data-based Counterfactual Data Augmentation (CDA) to fine-tune models and reduce bias. \citet{sun-etal-2019-mitigating} used debiasing methods in data manipulation, gender tagging for machine translation (MT), and bias fine-tuning, targeting FPED and FNED reductions via gender-name swapping. \citet{xie2023countergap} focused on gender bias evaluation through coreference resolution, using the Counter-GAP dataset from fictional texts, and demonstrated the effectiveness of name-based counterfactual data augmentation in mitigating bias over anonymization-based methods. 

\citet{masoudian-etal-2024-effective} proposed ConGater, a controllable gating mechanism to mitigate bias in classification and retrieval tasks by blending biased and debiased representations, with a loss function combining task performance and bias regularization to balance fairness and utility. \citet{igbaria-belinkov-2024-learning} proposed SimReg, a similarity-based regularization method that mitigates dataset bias by combining task loss (cross-entropy) with a cosine similarity loss that aligns representations of the biased model and an unbiased reference model. \citet{patel2024improving} and \citet{venugopal2024comprehensive} mitigated bias in classification tasks by combining cross-entropy loss with a Kullback–Leibler divergence term between predicted and uniform distributions, encouraging fairer and less biased model outputs through balanced prediction distributions.

\subsection{Bias Detection in Bangla}
There has been some recent research works on bias detection in Bangla. \citet{sadhu202empirical} investigates bias in Bangla by creating a dataset for intrinsic bias measurement and adapting existing methods to examine how varying context length impacts bias metrics. Both \citet{sadhu2024social} and \citet{sadhu2024empirical} focused on bias in large language models (LLMs) for Bangla, but with different scopes. \citet{sadhu2024social} examined both gender and religious biases in LLM outputs, introducing probing techniques to detect these biases, while \citet{sadhu2024empirical} concentrate on gendered emotion attribution, using a zero-shot learning approach to explore emotional stereotypes linked to gender in Bangla. However, all these works focused on specific types of bias \text{--} gender, religious, and emotional and the methodologies were used for only bias detection.

\section{Dataset Construction}

Due to the scarcity of Bangla datasets for gender bias studies, we curated and transformed existing task-specific Bangla datasets by selecting samples that contain male or female gender terms. Following established practices in LLM gender bias research in English \cite{kotek2023gender}, we then automatically generated new counterfactual samples by altering these gender-specific terms and names. Afterward, we separated male-centric and corresponding female-centric text samples from the original and counterfactual samples.

To identify gender terms, we developed a dictionary of 573 gendered word pairs in Bangla, sourced from the internet and grammar books. This dictionary features pairs like ``{\bng echel}'' (son) and ``{\bng emey}'' (daughter), and accounts for words with multiple feminine forms, such as ``{\bng bha{I}}'' (brother) with ``{\bng eban}'' (sister) and ``{\bng bhaib}'' (sister-in-law). We used this dictionary to generate samples by swapping the original gendered terms with their counterparts, ensuring all possible alternatives were considered. 
\begin{table}[h!]
\centering
\resizebox{0.7\columnwidth}{!}{%
\begin{tabular}{c|ccc}\toprule
\textbf{Task}                        & \textbf{Class}    & \textbf{Train} & \textbf{Test} \\ \midrule
\multirow{2}{*}{Sentiment}  & Positive & 3094  & 788  \\ 
                            & Negative & 788   & 216  \\ \midrule
\multirow{2}{*}{Sarcasm}    & Positive & 4250  & 1033 \\
                            & Negative & 3150  & 817  \\ \midrule
\multirow{2}{*}{Toxicity}   & Positive & 2840  & 746  \\ 
                            & Negative & 2802  & 664  \\ \midrule
\multirow{2}{*}{HateSpeech} & Positive & 4081  & 1152 \\ 
                            & Negative & 3921  & 848  \\ \bottomrule
\end{tabular}%
}
\caption{Data Distribution}
\label{tab:data-distribution}
\end{table}
For identifying names, we applied Named Entity Recognition (NER) and then replaced using a predefined set of masculine and feminine names based on the gender term present in each sample. Finally, human evaluators reviewed the automatically generated samples to address missed or unconverted gender terms and to correct errors from the NER model, such as missed names or incorrect identification of non-name words as names. We have included descriptions of the existing datasets, a detailed overview of our dataset construction process, and relevant dataset statistics in Appendix \ref{app:B}. Appendix \ref{app:C} lists sets of masculine and feminine names while Appendix \ref{app:GT} contains all the collected gendered terms. In addition, we provide a few sample entries from each dataset in Table~\ref{datasetExample} of Appendix~\ref{AppendixG}.

The data distribution for our experiments is presented in Table~\ref{tab:data-distribution}. We employed an 80-20 stratified sampling strategy for each classification task to split the data into training and testing sets. The names in the test set were manually replaced using a different distribution of predefined names, aiming to improve the model's generalization to unseen names. From the training data, 20\% was further set aside as a validation set to monitor the progress of debiasing. The training set was used to apply the debiasing strategies, while the test set was reserved for detecting and evaluating the effectiveness of the mitigation methods.

\section{Methodology}

Bangla poses distinctive challenges that exacerbate prediction mismatches in classification tasks involving gendered words, which are a root cause of gender bias, as the model may predict one version correctly while misclassifying the gender-swapped counterpart, leading to unequal treatment of male- and female-centric inputs. Unlike high-resource languages such as English, Bangla lacks grammatical gender but encodes gender implicitly through social roles, kinship terms, and names. For example, terms like {\bng bha{I}} (brother) and {\bng eban} (sister) suggest gender through semantics rather than syntax. This implicit encoding leads to models misclassifying counterfactual or gender-swapped inputs due to a lack of explicit cues. Moreover, polysemous gendered terms such as {\bng dada}, which may refer to “brother” or “grandfather,” create contextual ambiguity that models often fail to resolve, resulting in inconsistent predictions across different gendered forms of the same meaning. 
\begin{figure*}[!t]
    \centering
    \includegraphics[width=1\linewidth]{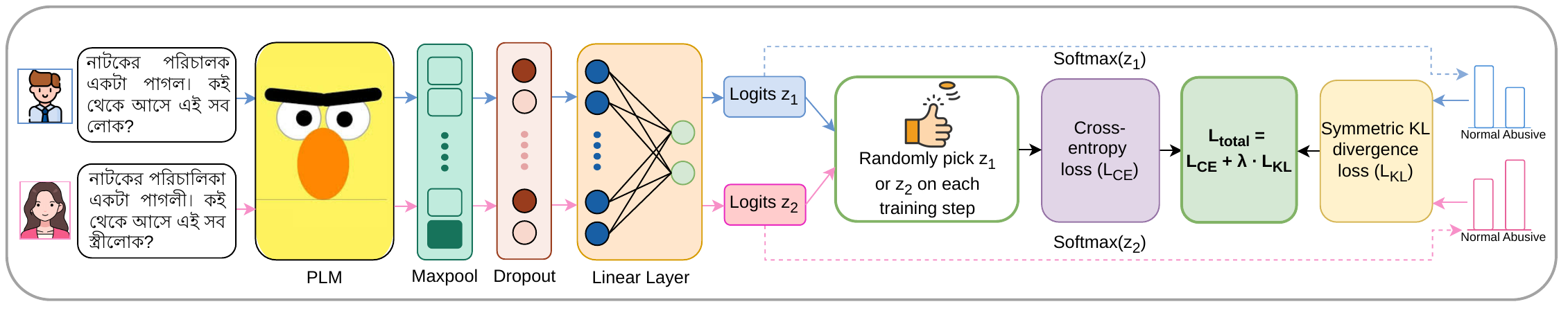}
    \caption{Overview of the proposed RandSymKL debiasing method.}
    \label{fig:RandSymKL}
\end{figure*}
Informal Bangla text further complicates classification due to widespread spelling variation and non-standard orthography (e.g., {\bng na{I}ma}, {\bng naiyma}, {\bng na{I}maH} referring to the same entity), making it difficult for token-based models to maintain prediction parity across gender pairs. Additionally, culturally entrenched terms like {\bng mailk} (owner) and {\bng grRiHNii} (housewife) carry gendered stereotypes that influence the model’s outputs even when gender is not overtly marked, skewing classification results. Finally, when multiple gendered terms co-occur in a sentence (e.g., “{\bng Aamar dada Aar mama bajaer igeyichl}”), transforming it into a feminine counterpart requires relational consistency. For example, mapping {\bng dada} to {\bng Thakurma}, {\bng idid}, or {\bng bhaib}, depending on context. These linguistic dependencies make it difficult for models that are not tailored to Bangla’s gender expression patterns to generate consistent predictions across male and female counterfactuals, thereby amplifying bias and reducing fairness in downstream tasks. 

To address these challenges, a mitigation strategy must go beyond surface-level substitutions and explicitly account for the contextual nature of gender in Bangla. Given these requirements, we design our proposed method, RandSymKL, as shown in figure \ref{fig:RandSymKL}, to operate at the output distribution level, aligning predictions across gender-swapped inputs without relying on token-level gender markers. It employs a stochastic fine-tuning strategy with a custom joint loss function. The objective was twofold: (1) minimize the cross-entropy loss for the classification task and (2) reduce the distance between the class probability distributions obtained from male-centric texts and corresponding female-centric texts via symmetric KL divergence loss.
% \begin{figure}[h!]
%     \centering
%     % \vspace{-27pt}
%     \includegraphics[width=1\linewidth]{images/method.pdf}
%     % \vspace{-40pt}
%     \caption{Overview of the proposed RandSymKL debiasing method.}
%     \label{fig:RandSymKL}
% \end{figure}
% % \vspace{-12pt}
\paragraph{Input Processing.}
At the start of the fine-tuning process, both the male-centric text and its corresponding female-centric text were fed into the model simultaneously. These paired inputs allowed the model to learn debiased representations by aligning the class probability distributions of male-centric ($P_{\text{male}}$) and corresponding female-centric texts ($P_{\text{female}}$), thereby encouraging the model to produce similar predictions across gendered text variants.

\paragraph{Embedding Extraction.}
Max-pooled embeddings were extracted for both the male-centric and the female-centric text from the model. Let the embeddings of the male and female texts be represented as $\mathbf{e}_1$ and $\mathbf{e}_2$, respectively. These embeddings were used in the joint loss computation to enforce similarity between the two.

\paragraph{Classification and Prediction.}
Each of the max-pooled embeddings, $\mathbf{e}_1$ and $\mathbf{e}_2$, was passed through a shared dropout layer, followed by a task-specific linear feed-forward layer with output dimensionality equal to the number of target classes. This produced the output logits $\mathbf{z}_1$ and $\mathbf{z}_2$, corresponding to the male-centric and female-centric inputs, respectively. The softmax-transformed outputs, $P_{\text{male}} = \text{softmax}(\mathbf{z}_1)$ and $P_{\text{female}} = \text{softmax}(\mathbf{z}_2)$, represent the predicted class probability distributions for each gendered input.

\paragraph{Randomized Loss Computation.}
To integrate fairness into training, a randomization strategy was employed. At each training step, either $\mathbf{z}_1$ or $\mathbf{z}_2$ was randomly selected and passed to the standard cross-entropy loss with respect to the true label. This randomness prevents the model from overfitting to a specific gendered phrasing during supervised training.

\paragraph{Symmetric KL Divergence Regularization.}  
To explicitly encourage consistency between gendered predictions, a symmetric KL divergence term was computed between $P_{\text{male}}$ and $P_{\text{female}}$:
\begin{equation}
\mathcal{L}_{\text{KL}} = \text{KL}(P_{\text{male}} \,\|\, P_{\text{female}}) + \text{KL}(P_{\text{female}} \,\|\, P_{\text{male}})
\end{equation}
This term penalizes asymmetric divergence in class probabilities between the gendered pairs, promoting alignment in their output distributions.
\paragraph{Joint Loss Optimization.}  
The final training objective combines the randomized cross-entropy loss $\mathcal{L}_{\text{CE}}$ and the symmetric KL divergence loss $\mathcal{L}_{\text{KL}}$ into a joint loss function:
\begin{equation}
\mathcal{L}_{\text{total}} = \mathcal{L}_{\text{CE}} + \lambda \cdot \mathcal{L}_{\text{KL}}
\end{equation}
Here, $\lambda$ is a hyperparameter that balances task accuracy and debiasing strength. This joint optimization guides the model toward fairness while preserving task-specific performance.

\paragraph{Training and Inference.}  
During training, the model optimized the joint loss function via backpropagation, simultaneously minimizing the classification error and aligning the output probability distributions of male-centric and corresponding female-centric inputs. This alignment, enforced through the symmetric KL divergence, encouraged the model to produce gender-invariant predictions. At inference time, only a single (original) input either male- or female-centric was required, as the model had already learned to generalize across gender variants during training. This eliminated the need to generate gender-swapped counterparts during inference, thereby reducing computational overhead.

\section{Experiments}

\paragraph{Models.} We employ four different Bangla PLMs, each fine-tuned for a specific downstream classification task. Two BERT-based \citet{devlin-etal-2019-bert} models are used for toxicity and hate speech detection, while the other two ELECTRA-based \citet{clark2020electra} models are utilized for sentiment and sarcasm classification. These task-specific fine-tuned models were selected for their robust performance and efficiency in Bangla NLP classification tasks. Detailed specifications of the models are provided in Appendix~\ref{app:D}.

\paragraph{Experimental Setup.}  For bias detection with the off-the-shelf inference (OSI) approach, we used default hyperparameters without additional fine-tuning. For bias mitigation, standard fine-tuning was applied. Texts were tokenized, passed through task-specific PLMs to generate token representations, and max-pooled for sentence representation. A dropout layer preceded the final linear layer. Across all experiments, we initially adopted consistent hyperparameters: a batch size of 4, a learning rate of $1e-4$, a dropout rate of 0.2, $\lambda = 1$, and the Adam optimizer. Each model was fine-tuned for 15 epochs. To achieve a better bias-variance trade-off, we monitored performance on the validation set and subsequently adjusted the dropout rate, followed by an additional 3 to 5 epochs of fine-tuning when necessary.

\paragraph{Evaluation Metrics.} 
We utilized several established evaluation metrics, FairScore ~\cite{qian-etal-2022-perturbation}, Accuracy Gap (AG) ~\cite{xie2023countergap, blodgett-etal-2020-language}, Group Fairness Metrics ~\cite{chen2024bias, shen-etal-2022-optimising} such as Absolute Statistical Parity Difference (SPD) and Absolute Equal Opportunity Difference (EOD), to quantify extrinsic gender bias. Detailed descriptions of the applied metrics are provided in appendix \ref{app:EM}.

\paragraph{Baselines for Bias Mitigation.} We experimented three baseline approaches for bias mitigation.

\noindent\textbf{(a) Fine tuning with only Original Data (FOD).} We fine-tuned task-specific models using the original, unaugmented dataset to assess whether fine-tuning alone could reduce bias compared to off-the-shelf inference. This setup offers a key comparison for evaluating the impact of fine-tuning on accuracy and bias mitigation against data augmentation or other interventions.

\noindent\textbf{(b) Token Masking (TM).} For each task, we replaced all names with a \verb|<Name>| token and gender terms with a \verb|<Gender>| token. Since these tokens were not in the tokenizer vocabulary of the pretrained models, we added them and adjusted the model's token embedding size accordingly before fine-tuning. This approach aims to eliminate bias by masking name and gender information entirely, preventing the model from relying on these identity specific cues during prediction.

\noindent\textbf{(c) Fine-tuning with both original and augmented samples (FOA).} 
For each task, we enhanced the original dataset by adding augmented data, which involved both male-centric and female-centric texts \cite{Lu2020}. This effectively doubled the dataset size from \textit{n} to \textit{2n} samples which we utilized for fine-tuning task-specific models.

\noindent\textbf{(d) Cosine Similarity-Based Debiasing (CSD).} This baseline adapted from \cite{igbaria-belinkov-2024-learning} employs a joint loss function ($J_{\text{loss}}$) that integrates classification and debiasing objectives. The classification component uses standard cross-entropy loss ($L_{\text{CE}}$), while the debiasing component ($L_{\text{GB}}$) encourages alignment between the embeddings of male- and female-centric text variants using cosine similarity. Specifically, the gender debiasing loss is defined as:
\begin{equation}
L_{\text{GB}} = 1 - \frac{\mathbf{e}_1 \cdot \mathbf{e}_2}{\|\mathbf{e}_1\| \|\mathbf{e}_2\|}
\end{equation}
where $\mathbf{e}_1$ and $\mathbf{e}_2$ are the max-pooled embeddings of the male- and female-centric inputs, respectively. The final loss combines both terms as:
\begin{equation}
J_{\text{loss}} = L_{\text{CE}} + \lambda L_{\text{GB}}
\end{equation}
Here, $\lambda$ is a hyperparameter that controls the trade-off between maintaining task performance and enforcing gender-invariant representations.

\noindent\textbf{(e) Non-Randomized Joint KLD Loss Minimization (NonRandSymKL\_M).}
This baseline adopts a joint loss formulation similar to our proposed method but without the randomized selection of cross-entropy inputs \cite{patel2024improving, venugopal2024comprehensive}. Specifically, the cross-entropy loss ($L_{\text{CE}}$) is always computed using the male-centric logits ($\mathbf{z}_1$), and combined with the symmetric KL divergence loss ($L_{\text{KL}}$) between the predicted distributions of male- and female-centric inputs. The underlying assumption of this baseline is that minimizing the KL divergence between $P_{\text{male}}$ and $P_{\text{female}}$ will implicitly align their class probability distributions. As a result, it is expected that using only the male-centric input for cross-entropy loss should suffice, since both gendered variants should converge to similar outputs through the KL regularization.

\noindent\textbf{(f) Averaged Joint KLD Loss Minimization (AvgSymKL\_MF).} 
This baseline modifies the NonRandSymKL\_M strategy by computing the average of the logits from the male- and female-centric inputs. Specifically, an averaged logits \( z_{\text{avg}} \) is obtained by taking the mean of the male and female logits. The cross-entropy loss is then computed using \( z_{\text{avg}} \) and the ground-truth label. A symmetric KL divergence term is also included to explicitly minimize the discrepancy between the male and female predictions. This approach assumes that the averaged logits provides a balanced training signal, promoting fairness across gendered variants while preserving task performance. 

\begin{table*}[h!]
\centering
\resizebox{\textwidth}{!}{%
\begin{tabular}{c|c|cc|cc|cc|cc|c}
\toprule
\rowcolor[HTML]{FFFFFF} 
\cellcolor[HTML]{FFFFFF} &
  \cellcolor[HTML]{FFFFFF} &
  \multicolumn{2}{c|}{\cellcolor[HTML]{FFFFFF}\textbf{Sentiment}} &
  \multicolumn{2}{c|}{\cellcolor[HTML]{FFFFFF}\textbf{Sarcasm}} &
  \multicolumn{2}{c|}{\cellcolor[HTML]{FFFFFF}\textbf{Toxicity}} &
  \multicolumn{2}{c|}{\cellcolor[HTML]{FFFFFF}\textbf{HateSpeech}} &
  \cellcolor[HTML]{FFFFFF} \\ 
\rowcolor[HTML]{FFFFFF} 
\multirow{-2}{*}{\cellcolor[HTML]{FFFFFF}\textbf{Approach}} &
  \multirow{-2}{*}{\cellcolor[HTML]{FFFFFF}\textbf{Data}} &
  \multicolumn{1}{c|}{\cellcolor[HTML]{FFFFFF}\textbf{Accuracy}} &
  \textbf{Mean} &
  \multicolumn{1}{c|}{\cellcolor[HTML]{FFFFFF}\textbf{Accuracy}} &
  \textbf{Mean} &
  \multicolumn{1}{c|}{\cellcolor[HTML]{FFFFFF}\textbf{Accuracy}} &
  \textbf{Mean} &
  \multicolumn{1}{c|}{\cellcolor[HTML]{FFFFFF}\textbf{Accuracy}} &
  \textbf{Mean} &
  \multirow{-2}{*}{\cellcolor[HTML]{FFFFFF}\textbf{\makecell{Overall Average\\Accuracy}}} \\ \midrule
 &
  Male Centric &
  \multicolumn{1}{c|}{26.09 {[}23.45, 29.22{]}} &
   &
  \multicolumn{1}{c|}{44.26 {[}41.97, 46.27{]}} &
   &
  \multicolumn{1}{c|}{71.91 {[}69.57, 74.11{]}} &
   &
  \multicolumn{1}{c|}{79.29 {[}77.29, 81.07{]}} &
   &
   \\ 
\multirow{-2}{*}{OSI} &
  Female Centric &
  \multicolumn{1}{c|}{29.78 {[}27.11, 32.68{]}} &
  \multirow{-2}{*}{27.94} &
  \multicolumn{1}{c|}{43.33 {[}41.13, 45.54{]}} &
  \multirow{-2}{*}{43.80} &
  \multicolumn{1}{c|}{69.18 {[}66.38, 71.81{]}} &
  \multirow{-2}{*}{70.55} &
  \multicolumn{1}{c|}{85.50 {[}84.05, 87.12{]}} &
  \multirow{-2}{*}{82.40} &
  \multirow{-2}{*}{56.17} \\ \midrule
 &
  Male Centric &
  \multicolumn{1}{c|}{96.82 {[}95.71, 97.83{]}} &
   &
  \multicolumn{1}{c|}{88.19 {[}86.78, 89.62{]}} &
   &
  \multicolumn{1}{c|}{91.25 {[}89.78, 92.62{]}} &
   &
  \multicolumn{1}{c|}{86.65 {[}95.07, 88.17{]}} &
   &
   \\ 
\multirow{-2}{*}{FOD} &
  Female Centric &
  \multicolumn{1}{c|}{95.10 {[}93.71, 96.44{]}} &
  \multirow{-2}{*}{\textbf{95.96}} &
  \multicolumn{1}{c|}{88.03 {[}86.67, 89.54{]}} &
  \multirow{-2}{*}{88.11} &
  \multicolumn{1}{c|}{89.63 {[}88.08, 91.11{]}} &
  \multirow{-2}{*}{90.44} &
  \multicolumn{1}{c|}{93.13 {[}91.95, 94.20{]}} &
  \multirow{-2}{*}{\textbf{89.89}} &
  \multirow{-2}{*}{\textbf{91.10}} \\ \midrule
TM &
  N/A &
  \multicolumn{1}{c|}{93.31 {[}91.80, 94.89{]}} &
  93.31 &
  \multicolumn{1}{c|}{87.06 {[}85.45, 88.59{]}} &
  87.06 &
  \multicolumn{1}{c|}{88.41 {[}86.73, 90.00{]}} &
  88.41 &
  \multicolumn{1}{c|}{81.05 {[}79.15, 82.77{]}} &
  81.05 &
  87.46 \\ \midrule
 &
  Male Centric &
  \multicolumn{1}{c|}{95.39 {[}94.12, 96.59{]}} &
   &
  \multicolumn{1}{c|}{88.92 {[}87.53, 90.29{]}} &
   &
  \multicolumn{1}{c|}{90.72 {[}89.29, 92.19{]}} &
   &
  \multicolumn{1}{c|}{87.17 {[}85.65, 88.75{]}} &
   &
   \\ 
\multirow{-2}{*}{FOA} &
  Female Centric &
  \multicolumn{1}{c|}{94.62 {[}93.09, 95.97{]}} &
  \multirow{-2}{*}{95.01} &
  \multicolumn{1}{c|}{88.85 {[}87.53, 90.21{]}} &
  \multirow{-2}{*}{88.89} &
  \multicolumn{1}{c|}{90.56 {[}89.21, 91.98{]}} &
  \multirow{-2}{*}{90.64} &
  \multicolumn{1}{c|}{87.47 {[}86.07, 88.85{]}} &
  \multirow{-2}{*}{87.32} &
  \multirow{-2}{*}{90.46} \\ \midrule
 &
  Male Centric &
  \multicolumn{1}{c|}{96.41 {[}95.15, 97.42{]}} &
   &
  \multicolumn{1}{c|}{88.28 {[}86.94, 89.7{]}} &
   &
  \multicolumn{1}{c|}{91.66 {[}90.24, 93.22{]}} &
   &
  \multicolumn{1}{c|}{88.19 {[}86.62, 89.77{]}} &
   &
   \\ 
\multirow{-2}{*}{CSD} &
  Female Centric &
  \multicolumn{1}{c|}{94.63 {[}92.99, 96.03{]}} &
  \multirow{-2}{*}{95.52} &
  \multicolumn{1}{c|}{87.62 {[}86.24, 89.02{]}} &
  \multirow{-2}{*}{87.95} &
  \multicolumn{1}{c|}{89.91 {[}88.26, 91.41{]}} &
  \multirow{-2}{*}{\textbf{90.79}} &
  \multicolumn{1}{c|}{87.97 {[}86.65, 89.45{]}} &
  \multirow{-2}{*}{88.08} &
  \multirow{-2}{*}{90.58} \\ \midrule
 &
  Male Centric &
  \multicolumn{1}{c|}{95.86 {[}94.48, 97.01{]}} &
   &
  \multicolumn{1}{c|}{88.13 {[}86.67, 89.51{]}} &
   &
  \multicolumn{1}{c|}{90.68 {[}89.18, 92.23{]}} &
   &
  \multicolumn{1}{c|}{88.77 {[}87.27, 90.20{]}} &
   &
   \\ 
\multirow{-2}{*}{NonRandSymKL\_M} &
  Female Centric &
  \multicolumn{1}{c|}{95.27 {[}93.91, 96.49{]}} &
  \multirow{-2}{*}{95.57} &
  \multicolumn{1}{c|}{87.75 {[}86.05, 88.94{]}} &
  \multirow{-2}{*}{87.94} &
  \multicolumn{1}{c|}{89.64 {[}88.18, 91.13{]}} &
  \multirow{-2}{*}{90.16} &
  \multicolumn{1}{c|}{88.69 {[}87.40, 89.95{]}} &
  \multirow{-2}{*}{88.73} &
  \multirow{-2}{*}{90.60} \\ \midrule
 &
  Male Centric &
  \multicolumn{1}{c|}{94.93 {[}94.51, 96.08{]}} &
   &
  \multicolumn{1}{c|}{89.23 {[}87.78, 90.62{]}} &
   &
  \multicolumn{1}{c|}{90.60 {[}89.08, 92.06{]}} &
   &
  \multicolumn{1}{c|}{88.57 {[}87.25, 90.00{]}} &
   &
   \\ 
\multirow{-2}{*}{AvgSymKL\_MF} &
  Female Centric &
  \multicolumn{1}{c|}{94.25 {[}92.73, 95.67{]}} &
  \multirow{-2}{*}{94.59} &
  \multicolumn{1}{c|}{88.64 {[}87.27, 90.08{]}} &
  \multirow{-2}{*}{\textbf{88.94}} &
  \multicolumn{1}{c|}{90.56 {[}89.15, 92.02{]}} &
  \multirow{-2}{*}{90.58} &
  \multicolumn{1}{c|}{88.55 {[}87.22, 89.85{]}} &
  \multirow{-2}{*}{88.56} &
  \multirow{-2}{*}{90.67} \\ \midrule
 &
  Male Centric &
  \multicolumn{1}{c|}{95.86 {[}94.64, 97.06{]}} &
   &
  \multicolumn{1}{c|}{88.07 {[}86.65, 89.54{]}} &
   &
  \multicolumn{1}{c|}{90.62 {[}89.08, 92.13{]}} &
   &
  \multicolumn{1}{c|}{88.39 {[}86.97, 89.80{]}} &
   &
   \\ 
\multirow{-2}{*}{RandSymKL} &
  Female Centric &
  \multicolumn{1}{c|}{95.46 {[}94.02, 96.65{]}} &
  \multirow{-2}{*}{95.66} &
  \multicolumn{1}{c|}{88.07 {[}86.70, 89.38{]}} &
  \multirow{-2}{*}{88.07} &
  \multicolumn{1}{c|}{90.14 {[}88.65, 91.63{]}} &
  \multirow{-2}{*}{90.38} &
  \multicolumn{1}{c|}{88.66 {[}87.35, 90.05{]}} &
  \multirow{-2}{*}{88.53} &
  \multirow{-2}{*}{90.66} \\ \bottomrule
\end{tabular}%
}
\caption{Classification accuracy evaluation across four tasks using various baseline and proposed debiasing methods. Each row reports the mean accuracy with 95\% confidence intervals over 500 runs for both male- and female-centric inputs. The `Mean' columns represent the average accuracy between male- and female-centric inputs for each task, while the `Average' column reflects the macro-average across all tasks.}
\label{tab:accuracy-table}
\end{table*}

\section{Results \& Analysis} 
To assess the mitigation performance, we answer the following research questions.

\paragraph{RQ1: How does the proposed bias mitigation technique affect accuracy?} A detailed accuracy comparison is presented in Table \ref{tab:accuracy-table}. Among the methods, FOD achieves the highest overall accuracy, followed closely by AvgSymKL\_MF. The proposed RandSymKL method attains a strong overall average accuracy of 90.66\%, which is only 0.44\% lower than FOD and 0.01\% lower than AvgSymKL\_MF. Excluding OSI, the TM approach exhibits the lowest accuracy at 87.46\%.
% \usepackage{graphicx}
\begin{table}[h!]
\centering
\resizebox{\columnwidth}{!}{%
\begin{tabular}{c|cccc|c}
\toprule
\textbf{Approach} & \textbf{Sentiment} & \textbf{Sarcasm} & \textbf{Toxicity} & \textbf{HateSpeech} & \textbf{Average AG} \\ \midrule
OSI              & 3.69          & 0.93          & 2.73          & 6.21          & 3.39          \\ 
FOD             & 1.72          & 0.16          & 1.62          & 6.48          & 2.50          \\ 
TM                & \textbf{0.00}      & \textbf{0.00}    & \textbf{0.00}     & \textbf{0.00}       & \textbf{0.00}       \\ 
FOA             & 0.77          & 0.07          & 0.16          & 0.30          & 0.32          \\ 
CSD             & 1.78          & 0.66          & 1.75          & 0.22          & 1.10          \\ 
NonRandSymKL\_M & 0.59          & 0.38          & 1.04          & 0.08          & 0.52          \\ 
AvgSymKL\_MF    & 0.68          & 0.59          & \textbf{0.04} & \textbf{0.02} & 0.33          \\ 
RandSymKL       & \textbf{0.40} & \textbf{0.00} & 0.48          & 0.27          & \textbf{0.29} \\ \bottomrule
\end{tabular}%
}
\caption{Accuracy Gap (AG) between male- and female-centric inputs across four 
classification tasks and different debiasing approaches. Lower values indicate
improved gender fairness. The Average AG column reflects the macro-average 
accuracy gap across all tasks. }
\label{tab:ag-table}
\end{table}
Regarding the accuracy gap between male- and female-centric inputs shown in Table \ref{tab:ag-table}, TM achieves a perfect fairness score with a gap of zero, but at the cost of lower accuracy. Conversely, RandSymKL demonstrates the lowest accuracy gap among all other approaches (0.29\%), followed by TM, outperforming other baselines in fairness. This indicates that RandSymKL strikes a superior balance between maintaining high accuracy and effectively mitigating gender bias compared to existing methods.

% \textbf{Key Takeaway: }Bias mitigation techniques can affect the accuracy of Bangla PLMs in different ways. TM is highly effective at eliminating bias but can lead to lower accuracy in some tasks, indicating a trade-off between bias reduction and model performance. The BA is effective for bias elimination compared to JLO, but JLO outperforms it in terms of overall average accuracy. JLO effectively reduces bias while maintaining high accuracy, making it a strong choice when both fairness and performance are essential.

\paragraph{RQ2: How effective is the proposed method for bias reduction?} Apart from accuracy gap, to measure gender bias we utilize additional fairness metrics including FairScore, Equal Opportunity Difference (EOD), and Statistical Parity Difference (SPD). As shown in Table~\ref{tab:fs-table}, FairScore evaluates the consistency of predictions between male- and female-centric texts. Among all methods, \textbf{TM} achieves a perfect FairScore of 0 across all tasks, indicating no prediction disparity; however, this comes at the cost of lower overall accuracy (see Table~\ref{tab:accuracy-table}). 

Excluding TM, the proposed \textbf{RandSymKL} method consistently outperforms all other baselines across all tasks with a substantial margin, achieving the lowest average FairScore of \textbf{1.69\%}. Although the absolute improvement over the strongest baseline (AvgSymKL\_MF: 2.30\%) appears modest, we further validate its reliability through statistical significance testing. Specifically, we apply McNemar's test \cite{mcnemar1947note, ogbuokiri2025cross} on paired consistency outcomes, where each male--female sentence pair is evaluated based on whether the model produces matching predictions. These paired outcomes directly correspond to the FairScore computation. The results show that RandSymKL achieves statistically significant reductions in gender inconsistency for Sentiment ($p = 0.029$) and Toxicity ($p = 0.021$). While Sarcasm ($p = 0.665$) and Hate Speech ($p = 0.880$) do not individually exhibit significance, a pooled analysis across all tasks yields a statistically significant overall improvement ($p = 0.012$).

% \usepackage{graphicx}
\begin{table}[h!]
\centering
\resizebox{\columnwidth}{!}{%
\begin{tabular}{c|cccc|c}
\toprule
\textbf{Approach} & \textbf{Sentiment} & \textbf{Sarcasm} & \textbf{Toxicity} & \textbf{HateSpeech} & \textbf{Average} \\ \midrule
OSI              & 25.67 & 2.43 & 37.38 & 22.75 & 22.06 \\
FOD             & 3.09  & 3.95 & 4.89  & 11.95 & 5.97  \\
TM                & \textbf{0.00}      & \textbf{0.00}    & \textbf{0.00}     & \textbf{0.00}       & \textbf{0.00}    \\
FOA             & 2.16  & 2.59 & 4.18  & 3.70  & 3.16  \\
CSD             & 2.16  & 3.19 & 4.33  & 3.55  & 3.31  \\
NonRandSymKL\_M & 1.65  & 1.89 & 3.05  & 2.65  & 2.31  \\
AvgSymKL\_MF    & 1.96  & 1.78 & 3.05  & 2.40  & 2.30  \\
RandSymKL         & \textbf{0.82}      & \textbf{1.57}    & \textbf{2.06}     & \textbf{2.30}       & \textbf{1.69}    \\ \bottomrule
\end{tabular}%
}
\caption{FairScore comparison across bias mitigation methods which illustrates percentage of prediction mismatches between male and corresponding female-centric texts. Lower values indicate improved gender fairness}
\label{tab:fs-table}
\end{table}
\begin{table*}[h!]
\centering
\resizebox{\textwidth}{!}{%
\begin{tabular}{c|cc|cc|cc|cc|c|c}
\toprule
\rowcolor[HTML]{FFFFFF} 
\cellcolor[HTML]{FFFFFF} &
  \multicolumn{2}{c|}{\cellcolor[HTML]{FFFFFF}\textbf{Sentiment}} &
  \multicolumn{2}{c|}{\cellcolor[HTML]{FFFFFF}\textbf{Sarcasm}} &
  \multicolumn{2}{c|}{\cellcolor[HTML]{FFFFFF}\textbf{Toxicity}} &
  \multicolumn{2}{c|}{\cellcolor[HTML]{FFFFFF}\textbf{HateSpeech}} &
  \cellcolor[HTML]{FFFFFF} &
  \cellcolor[HTML]{FFFFFF} \\ 
\rowcolor[HTML]{FFFFFF} 
\multirow{-2}{*}{\cellcolor[HTML]{FFFFFF}\textbf{Approach}} &
  \multicolumn{1}{c}{\cellcolor[HTML]{FFFFFF}\textbf{EOD}} &
  \textbf{SPD} &
  \multicolumn{1}{c}{\cellcolor[HTML]{FFFFFF}\textbf{EOD}} &
  \textbf{SPD} &
  \multicolumn{1}{c}{\cellcolor[HTML]{FFFFFF}\textbf{EOD}} &
  \textbf{SPD} &
  \multicolumn{1}{c}{\cellcolor[HTML]{FFFFFF}\textbf{EOD}} &
  \textbf{SPD} &
  \multirow{-2}{*}{\cellcolor[HTML]{FFFFFF}\textbf{Average EOD}} &
  \multirow{-2}{*}{\cellcolor[HTML]{FFFFFF}\textbf{Average SPD}} \\ \midrule
OSI &
  \multicolumn{1}{c}{0.036} &
  0.035 &
  \multicolumn{1}{c}{0.003} &
  0.004 &
  \multicolumn{1}{c}{0.054} &
  0.137 &
  \multicolumn{1}{c}{0.086} &
  0.110 &
  0.045 &
  0.072 \\ 
FOD &
  \multicolumn{1}{c}{0.016} &
  0.015 &
  \multicolumn{1}{c}{0.003} &
  0.004 &
  \multicolumn{1}{c}{0.006} &
  0.006 &
  \multicolumn{1}{c}{0.071} &
  0.077 &
  0.024 &
  0.026 \\ 
TM &
  \multicolumn{1}{c}{\textbf{0.000}} &
  \textbf{0.000} &
  \multicolumn{1}{c}{\textbf{0.000}} &
  \textbf{0.000} &
  \multicolumn{1}{c}{\textbf{0.000}} &
  \textbf{0.000} &
  \multicolumn{1}{c}{\textbf{0.000}} &
  \textbf{0.000} &
  \textbf{0.000} &
  \textbf{0.000} \\ 
FOA &
  \multicolumn{1}{c}{0.003} &
  0.004 &
  \multicolumn{1}{c}{0.003} &
  0.004 &
  \multicolumn{1}{c}{0.004} &
  0.005 &
  \multicolumn{1}{c}{0.003} &
  0.004 &
  0.003 &
  0.004 \\ 
CSD &
  \multicolumn{1}{c}{0.004} &
  0.009 &
  \multicolumn{1}{c}{0.005} &
  0.005 &
  \multicolumn{1}{c}{0.004} &
  0.012 &
  \multicolumn{1}{c}{0.003} &
  0.004 &
  0.004 &
  0.008 \\ 
NonRandSymKL\_M &
  \multicolumn{1}{c}{0.003} &
  0.010 &
  \multicolumn{1}{c}{0.005} &
  0.004 &
  \multicolumn{1}{c}{0.006} &
  0.004 &
  \multicolumn{1}{c}{0.003} &
  0.006 &
  0.004 &
  0.006 \\ 
AvgSymKL\_MF &
  \multicolumn{1}{c}{0.003} &
  0.007 &
  \multicolumn{1}{c}{0.003} &
  0.002 &
  \multicolumn{1}{c}{0.003} &
  0.004 &
  \multicolumn{1}{c}{0.002} &
  0.003 &
  0.003 &
  0.004 \\ 
RandSymKL &
  \multicolumn{1}{c}{\textbf{0.002}} &
  \textbf{0.003} &
  \multicolumn{1}{c}{\textbf{0.002}} &
  \textbf{0.002} &
  \multicolumn{1}{c}{\textbf{0.002}} &
  \textbf{0.003} &
  \multicolumn{1}{c}{\textbf{0.002}} &
  \textbf{0.003} &
  \textbf{0.002} &
  \textbf{0.003} \\ \bottomrule
\end{tabular}%
}
\caption{Absolute Equal Opportunity Difference (EOD), and Absolute Statistical Parity Difference (SPD) scores for various fairness intervention approaches. Each method reports the mean absolute EOD, and SPD over 500 runs (95\% bootstrap confidence interval (lower and upper bounds) can be found in Appendix~\ref{app:spd_eodapp} ). Lower values indicate greater fairness.}
\label{tab:spd-eod-table}
\end{table*}
A similar trend is observed in terms of EOD and SPD as presented in Table~\ref{tab:spd-eod-table}. While TM again achieves ideal fairness with zero EOD and SPD, \textbf{RandSymKL} records the next best performance with the lowest non-zero average EOD of \textbf{0.002} and SPD of \textbf{0.003}. Moreover, RandSymKL not only maintains fairness but also preserves high accuracy (see Table~\ref{tab:accuracy-table}), making it the most balanced approach in terms of both performance and bias mitigation. These results demonstrate the robustness and generalizability of the proposed method across different tasks and fairness metrics.

While \textbf{RandSymKL} substantially mitigates gender bias, we further explore the source of remaining prediction inconsistencies. As discussed in Appendix~\ref{app:rqi}, these appear to arise from subtle sensitivities in the final classification layer rather than significant representational divergences.

\paragraph{RQ3: Does minimizing FairScore lead to consistent improvements across standard group fairness metrics?}
Our proposed approach aims to align the output probability distributions between male- and female-centric inputs using a KL-divergence loss, which effectively penalizes prediction inconsistencies across gender-altered pairs. This regularization is directly analogous to minimizing FairScore, which measures the percentage of prediction mismatches across gender transformations. Our empirical findings reveal that optimizing this distributional alignment not only reduces FairScore but also consistently improves traditional group fairness metrics—demonstrating lower AG, EOD, and SPD across all evaluated tasks.

\paragraph{RQ4: What types of residual gender-related prediction errors persist after applying RandSymKL?} To answer this, we conducted a qualitative analysis of prediction mismatches in gender-swapped Bangla sentence pairs after debiasing. We identified four recurring error categories:

(1) Lexical stereotype bias: Certain male-centric phrases were more accurately classified due to societal stereotypes encoded in training data. For example, the sentence {\bng bha{I} khuish Hlam na} (I am not happy, brother) was correctly classified as negative sentiment, whereas its gender-swapped version {\bng Aapha khuish Hlam na} (I am not happy, sister) was misclassified, indicating a skewed model association between emotion and male terms.

(2) Morphological and orthographic rarity: Feminine forms that are rare or morphologically irregular were more prone to misclassification. For instance, the phrase {\bng kThiin guru} (Great boss) was predicted as positive, while the female form {\bng kThiin gur/dhii} (Great boss lady) was misclassified, likely due to lower frequency or unfamiliarity of the token {\bng gur/dhii}.

(3) Toxicity detection asymmetry: Toxicity was more reliably flagged in male-centric sentences. For example, {\bng bha{I} jy mN/Dl Hela AkrRtg/Y} (Brother Joy Mondol is ungrateful) was classified as toxic, but {\bng eban imm Hela AkrRtg/Y} (Sister Mim is ungrateful) was not, despite identical semantic framing.

(4) Name familiarity bias: The model showed better accuracy for popular male names like {\bng inesha} or {\bng {I}mdadul} compared to less common female names such as {\bng  nujHat} or {\bng munmun} affecting performance across sentiment and sarcasm tasks.

These examples highlight that despite improved fairness, RandSymKL cannot fully mitigate biases rooted in language use patterns, data distributions, and sociocultural associations in Bangla.

% \paragraph{Does confidence alignment between gendered input pairs correlate with improved fairness and predictive consistency across confidence levels?}

% \section{Future Work}
% While this study provides valuable insights into gender bias in Bangla pretrained language models and evaluates several mitigation techniques, there is considerable scope for further research, particularly concerning the trade-off between bias reduction and accuracy. One of the main challenges identified in this study is balancing bias mitigation with maintaining model performance. Some techniques, like Token Masking (TM), effectively reduce bias to zero but at the cost of decreased accuracy in certain tasks. This trade-off highlights a critical area for future work: understanding the underlying mechanisms that cause accuracy to decline as bias is reduced. Further research should investigate the intricate relationship between bias mitigation and model accuracy, exploring why certain techniques perform better in balancing these two aspects than others. This could involve analyzing how different types of gender biases are encoded in pre-trained models and how specific mitigation strategies impact model weights and feature representations. Additionally, future research could focus on developing and refining more sophisticated bias mitigation techniques that can adapt to various contexts and tasks. For instance, exploring adaptive learning algorithms that can dynamically adjust to different types of biases encountered during training could provide more robust solutions. 

\section{Conclusion}
This study reveals the presence of extrinsic gender bias in Bangla pretrained language models across several classification tasks. Our bias mitigation approach effectively reduced bias by consistently outperforming other baselines across different tasks while maintaining competitive accuracy, indicating its potential for wider use in NLP tasks. These findings highlight the need to address gender bias to ensure fair outcomes in NLP applications, especially for under-resourced languages like Bangla.

\section{Limitations}
During dataset transformation, we swapped male-centric names with female-centric ones and vice versa using an NER model to identify names in the original texts. However, the accuracy of the NER model, particularly in Bangla, might be limited and may require improvement. Moreover, our approach assumes binary gender categories, excluding non-binary identities, which restricts the scope of bias detection. We did not use large language models (LLMs) in this study, which could limit the relevance of our findings for applications involving state-of-the-art models. In future work, our benchmark datasets could be used to evaluate the behavior of LLMs in zero-shot settings, providing further insight into their gender bias tendencies. Additionally, our method could be applied to other low-resource or morphologically rich languages to assess its effectiveness across linguistic and cultural contexts. Furthermore, we used consistent hyperparameters across all mitigation approaches, but altering these could affect the outcomes. Finally, the datasets may lack sufficient diversity and balance, impacting the generalizability of the results.

\section{Ethical Considerations}
Our research focuses on binary gender bias due to the structure of existing datasets and literature. We recognize the importance of non-binary and gender-fluid identities and encourage future work to include these dimensions for a more comprehensive approach to bias mitigation. Our datasets contain toxic and hate speech content, which may be offensive to some. We retained these data points to ensure that our models are exposed to realistic, harmful language, emphasizing the importance of bias mitigation in language models that interact with such content.

\bibliography{bibliography}

\appendix
\section*{Appendix}
\label{sec:appendix}

\section{Bangla Gender Bias Dataset}\label{app:B}
To the best of our knowledge, there are currently no datasets specially designed to identify extrinsic gender bias in Bangla classification tasks. Therefore, to explore gender bias, we curated four task specific datasets. 

\subsection{Dataset Collection}
We collected datasets from four different sources along with the original labels.\\

\noindent\textbf{(a) Sentiment Detection.} We utilized the dataset provided by \cite{sazzed2020cross}, which consists of 11,807 texts. After keeping the samples that have gender terms, we got 4,886 instances.\\

\noindent\textbf{(b) Sarcasm Detection.} We utilized the Ben-Sarc dataset provided by \cite{lora2023transformer}, which consists of a total of 25,636 Bangla social media comments where half of the data are sarcastic and rest half are non-sarcastic. This dataset had 9,250 texts with gender terms.\\

\noindent\textbf{(c) Toxicity Detection.} For toxicity detection, we used a multi labeled dataset provided by \cite{belal2023interpretable}, which has 16,073 sentences. Among them, 7,585 comments are labeled as non-toxic and 8,488 comments are labeled as toxic. As we needed only those texts that have gender terms, we finally gathered 7,052 texts for our study.\\

\noindent\textbf{(d) Hate-Speech Detection.} For hate speech detection, we used a dataset from \cite{romim2021hate}, that has a total of 30,000 sentences. Among them, 10,000 are hate speech and 20,000 are non-hate speech comments. After discarding non-gendered comments, we have a total of 10,002 comments.

\begin{table} [t]
    \centering
    \begin{tabular}{p{3.5cm}p{3.5cm}}
     \hline
     \textbf{Original Sentence} & \textbf{Converted Sentence}  \\
    \hline
     \textbf {[B]}{ \bng iThk belechn \colorbox{orange!50}{dada} Aamar men Hy Unar mathay smsYa Aaech} & \textbf {[B]}{ \bng iThk belechn \colorbox{green!50}{ebouid} Aamar men Hy Unar mathay smsYa Aaech}\\
      \textbf {[E]} Rightly said \colorbox{orange!50}{brother} I think he has a problem in his head & \textbf {[E]} Rightly said \colorbox{green!50}{sister-in-law} I think he has a problem in his head\\ \hline
       \textbf {[B]}{ \bng iThk belechn \colorbox{orange!50}{dada} Aamar men Hy Unar mathay smsYa Aaech} & \textbf {[B]}{ \bng iThk belechn \colorbox{green!50}{dadii} Aamar men Hy Unar mathay smsYa Aaech}\\ 
       \textbf {[E]} Rightly said \colorbox{orange!50}{grandfather} I think he has a problem in his head & \textbf {[E]} Rightly said \colorbox{green!40}{grandmother} I think he has a problem in his head\\ \hline
       \textbf {[B]}{ \bng iThk belechn \colorbox{orange!50}{dada} Aamar men Hy Unar mathay smsYa Aaech} & \textbf {[B]}{ \bng iThk belechn \colorbox{green!50}{idid} Aamar men Hy Unar mathay smsYa Aaech}\\ 
       \textbf {[E]} Rightly said \colorbox{orange!50}{brother} I think he has a problem in his head & \textbf {[E]} Rightly said \colorbox{green!50}{sister} I think he has a problem in his head\\ \hline
      
    \end{tabular}
    \caption{Handling generated samples when a gender term in the original sample has multiple meanings. Here \textbf {B} denotes the text in Bangla and \textbf {E} denotes the corresponding English translation.\colorbox{orange!50}{Orange} denotes gender terms with multiple meanings, while \colorbox{green!40}{green} indicates the counterparts of that term.}
    \label{convertedSentence}
\end{table}

\subsection{Bias Dataset Construction}

\noindent\textbf {(a) Dictionary Construction.} We start by creating a dictionary of gendered word pairs in Bangla which is utilized to distinguish gender-specific terms in texts. We collected these words from the internet and also from a few Bangla grammar books. This dictionary consists of 573 pairs of masculine and feminine words. All the gender terms can be found in Appendix~\ref{app:GT}. For instance, “{\bng echel}” (son) and “{\bng emey}” (daughter) are included in a pair as key. In cases where certain words had multiple potential replacements, all the words were included. For example, the word “{\bng bha{I}}” (brother) has two pairs with its feminine counterparts “{\bng eban}” (sister) and “{\bng bhaib}” (sister-in-law). In Bangla, the same words can be spelled with different letters, leading to various spellings. To address this, we incorporated spelling variations into our dictionary. For instance, male terms such as "{\bng ibhkharii}" and "{\bng ibhkhair}" correspond to female terms like "{\bng ibhkhairnii}", reflecting spelling differences. Similarly, the male term "{\bng nana}" relates to female terms like "{\bng nain}" and "{\bng nanii}".\\

\noindent\textbf {(b) Identify Gender Terms.} Initially, we tokenized each word of each text in the collected datasets. We then examined each token to see if it matched any words in the dictionary. If a match was found, we retained the text for containing a gender term; otherwise, we discarded it due to the absence of gender terms. This method resulted in four datasets containing texts with gendered terms across the four distinct tasks.\\

\noindent\textbf {(c) Gender Swapping.} To identify gender bias, we need a pair of samples. The first text in the pair is the original text with gender-specific terms, which we automatically convert by swapping these terms with their counterparts. This process starts by tokenizing each word of each text in the dataset. We then check each token against a dictionary to find matches. When a match is found, we replace it with its counterpart, leaving unmatched tokens unchanged. For example, the word "{\bng s/tRii}" (wife) would be replaced with "{\bng sWamii}" (husband). Some masculine or feminine words may have multiple suitable counterparts already included in the dictionary. For instance, the masculine word "{\bng dada}" can mean both brother and grandfather, so its feminine counterparts could be "{\bng dadii}" (grandmother), "{\bng idid}" (sister), and "{\bng ebouid}" (sister-in-law). For such cases, we generate samples with all possible counterparts for a single original sample. Few examples of these text transformations are shown in Table~\ref{convertedSentence}.\\

\noindent\textbf {(d) Name Replacement.} After swapping gender-specific terms, some texts still contained names. To ensure consistency between names and gender in the converted texts, we swapped names based on their gender relevance using a Named Entity Recognition (NER) model\footnote{\url{https://sparknlp.org/2021/02/10/bengaliner_cc_300d_bn.html}} which is trained on data provided by \cite{8944804}. The NER model tokenizes the text and identifies various entities such as person names, locations, date and time, organization names, etc., but we focused solely on person names. When a name is detected, we determine its gender and replace it with a corresponding name from the predefined set of male or female names. Texts without person names remained unchanged. This method maintained gender and name consistency across the datasets, resulting in equal numbers of original and transformed texts. A complete example of the gender and name swapping process is provided in Table~\ref{CompleteExample}.

\begin{table}[h!]
     \centering
    \begin{tabular}{p{1.6cm}p{5.3cm}}
    \hline
    Original Sentence & \textbf {[B]}{ \bng itsha \colorbox{red!70}{Aapu} khub sun/dr kibta ilkhet paern.}
    \textbf {[E]} \colorbox{red!70}{Sister} Tisha can write very beautiful poems. \\ 
    \hline
    Gender Term Converted Sentence & \textbf {[B]}{ \bng itsha \colorbox{yellow!70} {bha{I}ya} khub sun/dr kibta ilkhet paern.}
    \textbf {[E]} \colorbox{yellow!70}{Brother} Tisha can write very beautiful poems.\\ 
    \hline
    Name Detection using NER model & \textbf {[B]}{ \bng \colorbox{orange!70}{itsha} bha{I}ya khub sun/dr kibta ilkhet paern.}
    \textbf {[E]} Brother \colorbox{orange!70}{Tisha} can write very beautiful poems.\\ 
    \hline
    Final Converted Sentence & \textbf {[B]} {\bng \colorbox{green!50}{Aashraph} bha{I}ya khub sun/dr kibta ilkhet paern.}
    \textbf {[E]} Brother \colorbox{green!50}{Ashraf} can write very beautiful poems.\\
    \hline
    \end{tabular}
    
    \caption{Example of the gender-name swapping process step by step. Here \textbf {B} denotes the text in Bangla and \textbf {E} denotes the corresponding English translation. \colorbox{red!70}{Red} denotes the gender term that we want to swap, to generate a new sample. \colorbox{yellow!70}{Yellow} indicates the counterpart of that gender term which has been swapped. \colorbox{orange!70}{Orange} indicates the name from the transformed sample that we have to change. \colorbox{green!50}{Green} indicates the replaced name with relevance to gender.}
    \label{CompleteExample}
\end{table}

\noindent\textbf {(e) Grouping Samples According to Gender.}
Now we have obtained original texts along with their corresponding counterfactual versions that reflect the opposite gender. To measure bias effectively, it was necessary to pair each male-centric text with its corresponding female-centric counterpart—regardless of whether it originated from the original or the counterfactual set. This pairing was accomplished through dictionary-based matching.

\subsection{Human Evaluation}
In this study, we recruited four annotators who are undergraduate students with prior experience in Bangla grammar, spelling, and Named Entity Recognition tasks. These annotators were chosen from a pool of ten candidates. Initially, all participants took an exam with 50 questions focused on gender term conversion, correcting spelling errors, and identifying named entities. The four students with the highest scores were selected as annotators. After each step of the dataset construction process, the annotators manually reviewed the changes and made corrections.

After converting the gender terms in the original texts, the annotators manually reviewed the transformed texts for each of the four datasets, with each annotator responsible for one dataset. During the review, they found some gender terms that had been missed and not converted to their corresponding male or female counterparts. The annotators corrected these errors and updated the dictionary with the missing terms. In total, 28 missing gender terms were identified across all datasets.

\begin{table*}
\centering
\resizebox{\textwidth}{!}{
\begin{tabular}{>{\centering\hspace{0pt}}m{0.138\linewidth}>{\centering\hspace{0pt}}m{0.123\linewidth}>{\centering\hspace{0pt}}m{0.146\linewidth}>{\centering\hspace{0pt}}m{0.123\linewidth}>{\centering\hspace{0pt}}m{0.121\linewidth}>{\centering\hspace{0pt}}m{0.121\linewidth}>{\centering\arraybackslash\hspace{0pt}}m{0.154\linewidth}} \toprule
\textbf{Task}     & \textbf{No. of }\par{}\textbf{original}\par{}\textbf{samples} & \textbf{No. of }\par{}\textbf{converted }\par{}\textbf{samples} & \textbf{No. of }\par{}\textbf{total }\par{}\textbf{samples} & \textbf{Average }\par{}\textbf{no. of }\par{}\textbf{word} & \textbf{Average}\par{}\textbf{no. of}\par{}\textbf{gender }\par{}\textbf{terms} & \textbf{Frequency}\par{}\textbf{of}\par{}\textbf{names}  \\ \midrule
Sentiment         & 4,886                                                         & 4,886                                                           & 9,772                                                       & 19.668                                                     & 1.500                                                                           & 4957                                                     \\ 
Sarcasm           & 9,250                                                         & 9,250                                                           & 18,500                                                     & 19.187                                                     & 1.470                                                                           & 4677                                                     \\ 
Toxicity          & 7,052                                                         & 7,052                                                           & 14,104                                                      & 18.345                                                     & 1.587                                                                           & 4695                                                     \\ 
Hate Speech & 10,002                                                      & 10,002                                                          & 20,004                                                      & 25.946                                                     & 2.106                                                                           & 5343                                                     \\ \bottomrule
\end{tabular}}
\caption{Dataset statistics for each classification task, including the number of original and converted samples, total sample count, average word count per sample, average number of gender-related terms, and the total count of names in the original samples. From original and converted samples, male and corresponding female or female and corresponding male centric samples were separated}
\label{Statistics2}
\end{table*}

Some of the datasets contained lengthy texts with multiple gender terms to detect. However, the annotators found that certain gender terms were not swapped due to the complexity of handling male terms like "{\bng bha{I}eyra}" and "{\bng jama{I}ek}", and female terms like "{\bng ebaenra}" and "{\bng bUek}", among others. Across all datasets, there were 397 instances of such cases. 

In Bangla, a word can have multiple meanings, making accurate conversion difficult. For example, the original sentence: "{\bng puruSh ten/tRr ishkar Hey ta{I} puruShra ejel EbNNG irmaen/D Aaech.}" should have been converted to: "{\bng narii ten/tRr ishkar Hey ta{I} nariira ejel EbNNG irmaen/D Aaech.}" However, it was incorrectly converted, resulting in nonsensical sentences like: "{\bng narii ten/tRr ishkar Hey ta{I} nariira ejelin EbNNG irmaen/D Aaech.}" or "{\bng narii ten/tRr ishkar Hey ta{I} nariira ejel bU EbNNG irmaen/D Aaech.}" This was not an appropriate context for changing gender terms. Similar issues occurred with words like "{\bng ja}" to "{\bng ed{O}r}" and "{\bng khan}" to "{\bng khanm}". These errors were corrected by the annotators, and there were 548 such cases across all datasets.

After applying Named Entity Recognition (NER) to the samples in our datasets, the annotators found that the model incorrectly flagged some irrelevant words as names, which were later excluded from the analysis. In Bangla, names can have multiple spellings, causing the model to miss names. For instance, "{\bng inesha}" and "{\bng inshu}" refer to the same person, but the model fails to detect them due to spelling variations. Also the model also struggled to detect names used as third-person singular possessives, like "{\bng emHjaibenr}" (Mehjabin’s), leaving some names undetected. The annotators corrected 749 such cases across all datasets.

% In some sentences, the model incorrectly selected gender terms along with the names. Since this was not our objective, these instances were checked manually, and only the names were retained.

\subsection{Dataset Statistics}
Table~\ref{Statistics2} presents the data distribution across all datasets. The equal number of samples between the original and transformed datasets ensures a fair comparison for evaluating bias. The average word count per sample offers insights into the typical length and complexity of the samples. Variations in the frequency of gender terms and name counts across tasks highlight the diversity of the datasets. 

\section{Male and Female Names}\label{app:C}
We list the male and female names used for transforming task specific datasets below. We collected these names provided by \cite{das2023toward}.\\
\noindent\textbf {Male names.} {\bng ishb, kair/tk, bruN, menaHr, pRbal, ramkumar, shiil, Ark, AirtR, inly, pRtiik, sn/t, pRan/t, nyn, Aab/dulLaH, Aab/dur, {I}mdadul, {I}Usuph, Aashraph, kamal, juliphkar, naijrul, shamsuddiin, Aaisr, Aaitkur, naiphs, taHimd, masud, sadman, AaHnaph}\\
\noindent\textbf {Female names.} {\bng lkKii, srsWtii, kaliitara, dur/ga, saibtRii, dmyn/tii, tptii, ibinta, srla, suisMta, AmrRta, sunn/da, AadrRta, ismn/tii, Antra, gulshan, ejaHra, jaHan, Aaeysha, nuurjaHan, saHana, Haibba, najinn, ra{I}sa, nujHat, ma{I}sha, pharHana, saidya, naHar, musiphka}\\
To change the names of the test set, the following name distribution is used.\\
\noindent\textbf {Male names.} {\bng joy, shaOn, topu, raikb, jbbar}\\
\noindent\textbf {Female names.} {\bng raimsa, imm, sumona, joba, sanijda}
\section{Model Descriptions} \label{app:D}

\noindent\textbf{(a) BanglaSARC \footnote{\url{https://huggingface.co/raquiba/sarcasm-detection-BanglaSARC}}.} An ELECTRA \cite{clark2020electra} variant which is fine-tuned on Bangla sarcasm data provided by \cite{apon2022banglasarc}. The model has an embedding size of 768, vocabulary size of 32000 and it has a total of 110M parameters.\\

\noindent\textbf{(b) Abusive-MuRIL\footnote{\url{https://huggingface.co/Hate-speech-CNERG/bengali-abusive-MuRIL}}.} MuRIL is a BERT \cite{devlin-etal-2019-bert} variant that has been pre-trained on 17 Indian languages, including their transliterated versions \cite{khanuja2021muril}. Abusive-Muril is a fine-tuned MuRIL model pretrained for abusive hate speech detection task provided by \cite{das2022data}. The model has 110 million parameters and assigns an embedding size of 768 to each token. It uses a vocabulary of size 197285.\\

\noindent\textbf{(c) BanglaBERT-Sentiment\footnote{\url{https://huggingface.co/ka05ar/banglabert-sentiment}}.} An ELECTRA variant which has been fine-tuned over a sentiment dataset provided by \cite{blp2023-overview-task2}. It has 110 million trainable parameter with 32000 vocabulary size and a 768 token embedding size.\\

\noindent\textbf{(d) one-for-all-toxicity\footnote{\url{https://huggingface.co/FredZhang7/one-for-all-toxicity-v3}}.} A BERT variant model\cite{FredZhangUBC_undated} which has been fine-tuned using toxi-text-3M dataset\footnote{\url{https://huggingface.co/datasets/FredZhang7/toxi-text-3M}} to detect 13 different types of toxicity.  This model has 110 million parameter, 768-dimensional token embedding with a vocabulary size of 119547. 

\section{Description of Evaluation Metrics}\label{app:EM}
\paragraph{FairScore Metric.}  
We adapt the FairScore metric from \cite{qian-etal-2022-perturbation} to quantify gender bias by measuring the prediction inconsistency between male-centric and corresponding female-centric inputs. For a classifier $f_C$ and a pair of semantically equivalent gendered inputs $(x_{\text{male}}, x_{\text{female}})$, we consider the model biased if $f_C(x_{\text{male}}) \ne f_C(x_{\text{female}})$. Given an evaluation set $X$ composed of such gendered pairs, we define the FairScore (FS) as:
\begin{equation}
\text{FS}(f_C, X) = \frac{|\{x \in X \,|\, f_C(x_{\text{male}}) \ne f_C(x_{\text{female}})\}|}{|X|}
\end{equation}
A higher FairScore indicates greater gender-based inconsistency in predictions, and thus, higher bias. In our setting, this metric captures extrinsic gender bias by reflecting the model's differential treatment of semantically identical inputs differing only in gendered terms.

\paragraph{Accuracy Gap Metric.}   We compute the \textit{Accuracy Gap} (AG) \cite{xie2023countergap, blodgett-etal-2020-language}, to measure disparities in predictive performance across gendered variants. For each pair of male-centric and female-centric inputs $(x_{\text{male}}, x_{\text{female}})$ with ground-truth label $y$, we compute the classification accuracy separately on each group:
\begin{equation}
\text{AG}(f_C, X) = \left| \text{Acc}_{\text{male}} - \text{Acc}_{\text{female}} \right|
\end{equation}
where $\text{Acc}_{\text{male}}$ and $\text{Acc}_{\text{female}}$ denote the classification accuracy on the male-centric and female-centric texts, respectively. A lower Accuracy Gap indicates that the model performs more uniformly across gendered inputs, reflecting improved fairness.

\paragraph{Group Fairness Metrics.}  To quantify group-level fairness, we report two standard metrics adapted from \cite{chen2024bias, shen-etal-2022-optimising}: \textit{Absolute Statistical Parity Difference (SPD)} and \textit{Absolute Equal Opportunity Difference (EOD)}. SPD measures the absolute difference in positive prediction rates between male- and female-centric inputs, regardless of the true label:

\begin{equation}
\text{SPD} = \left| \Pr(f_C(x_{\text{male}}) = 1) - \Pr(f_C(x_{\text{female}}) = 1) \right|
\end{equation}
EOD, on the other hand, measures the absolute difference in true positive rates between groups, considering only the samples with a positive ground-truth label. It is defined as:
\begin{equation}
\text{EOD} = \left| \text{TPR}_{\text{male}} - \text{TPR}_{\text{female}} \right|
\end{equation}
where \( \text{TPR}_{\text{male}} \) and \( \text{TPR}_{\text{female}} \) denote the true positive rates for the male and female groups, respectively, computed as:
\begin{equation}
\text{TPR}_{\text{male}} = \Pr(f_C(x_{\text{male}}) = 1 \mid y = 1)
\end{equation}
\begin{equation}
\text{TPR}_{\text{female}} = \Pr(f_C(x_{\text{female}}) = 1 \mid y = 1)
\end{equation}
Lower values for both SPD and EOD indicate improved fairness by ensuring that the classifier treats male- and female-centric inputs more equally in terms of prediction frequency and correctness.

\section{Frequency of Bias}\label{AppendixBias}
Table \ref{bias_table} presents the exact number of mismatched predictions by task specific models between the male-centric samples and corresponding female-centric samples across different mitigation techniques for the four tasks.

% \usepackage{graphicx}
\begin{table}[h!]
\centering
\resizebox{0.5\textwidth}{!}{%
\begin{tabular}{ccccc}
\toprule
\rowcolor[HTML]{FFFFFF} 
\textbf{Approach} & \textbf{Sentiment} & \textbf{Sarcasm} & \textbf{Toxicity} & \textbf{HateSpeech} \\ \midrule
OSI              & 249 & 45          & 527 & 455 \\
FOD             & 30  & 73          & 69  & 239 \\ 
TM              & 0   & 0           & 0   & 0   \\ 
FOA             & 21  & 48          & 59  & 74  \\ 
CSD             & 21  & 59          & 61  & 71  \\ 
NonRandSymKL\_M & 16  & \textbf{35} & 43  & 53  \\ 
AvgSymKL\_MF    & 19  & 33          & 43  & 48  \\ 
RandSymKL       & 8   & 29          & 29  & 46  \\ \bottomrule
\end{tabular}%
}
\caption{Absolute prediction mismatches between male and female centric texts across bias mitigation methods. Lower mismatch counts suggest improved fairness and gender-invariant behavior in model predictions.}
\label{bias_table}
\end{table}

\begin{figure}[h!]
    \centering
    \includegraphics[width=\columnwidth, trim={0 7cm 0 7cm}]{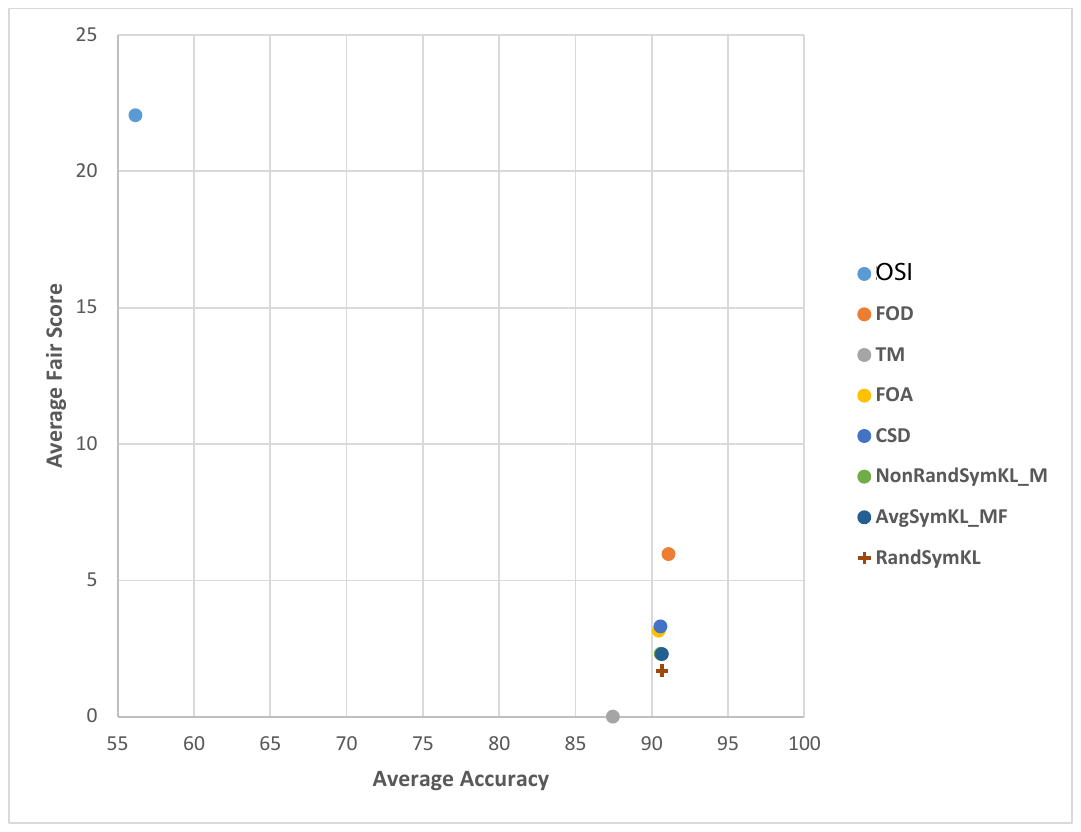}
    \caption{Bias vs Accuracy trade-off of different mitigation approaches.}
    \label{fig:bva}
\end{figure}

\begin{figure*}[h!]
		\centering
		\begin{subfigure}[b]{0.45\textwidth}
		    \centering
			\includegraphics[scale=0.4]{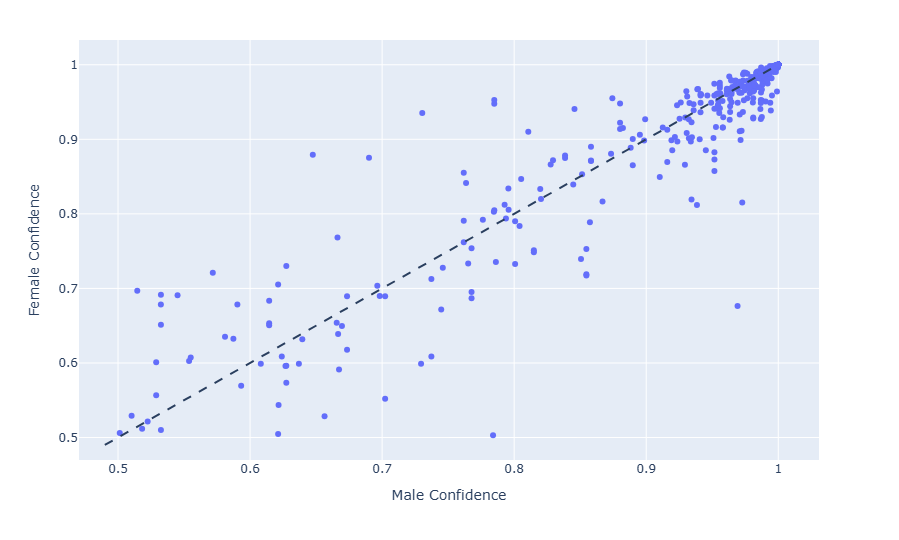}
                 \subcaption{}
                \label{fig:Usa_before}
		\end{subfigure}
            \begin{subfigure}[b]{.45\textwidth}
		    \centering
			\includegraphics[scale=0.4]{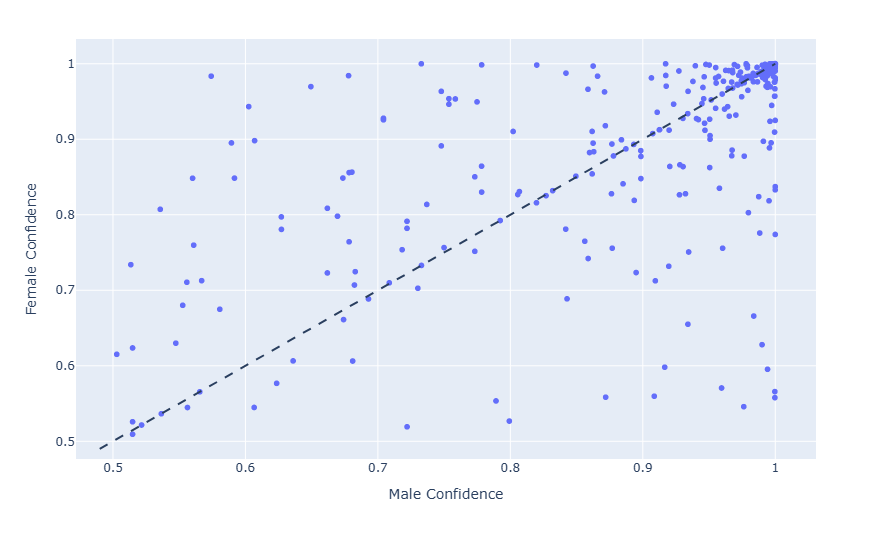}
                \subcaption{}
                \label{fig:Comp_before}
		\end{subfigure}\\
        
        \begin{subfigure}[b]{.45\textwidth}
		    \centering
			\includegraphics[scale=0.4]{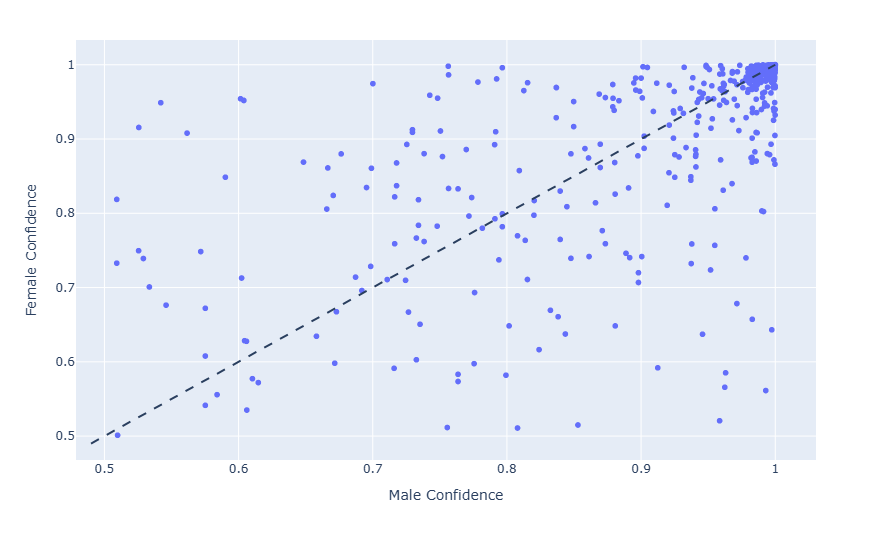}
			\subcaption{}
                \label{fig:Bug_before}
		\end{subfigure}  
        \begin{subfigure}[b]{.45\textwidth}
		    \centering
			\includegraphics[scale=0.4]{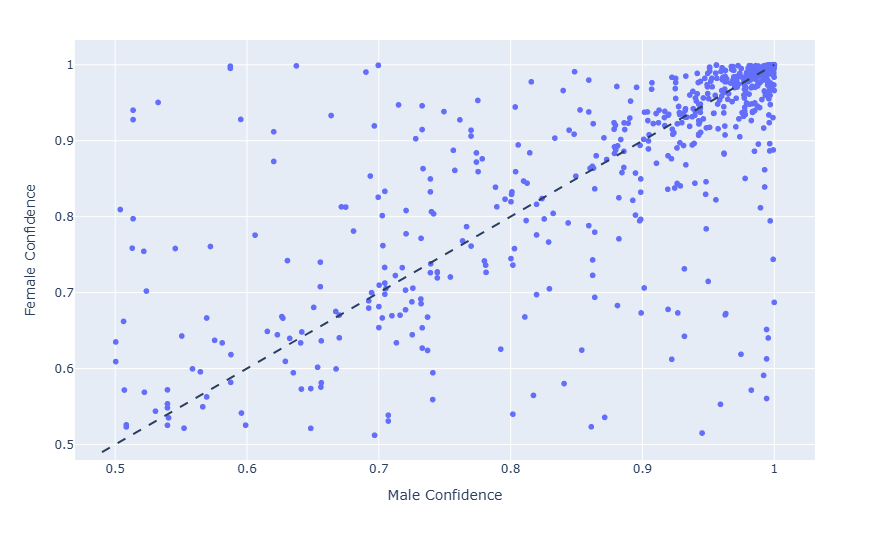}
                 \subcaption{}
                \label{fig:Usa_after}
		\end{subfigure}
		
		\caption{Confidence score visualization for male and corresponding female-centric text across a) sentiment analysis, b) sarcasm detection, c) toxicity detection, and d) hate speech detection}
		\label{fig:mfc}
\end{figure*}

\begin{figure*}[h!]
		\centering
		\begin{subfigure}[b]{0.45\textwidth}
		    \centering
			\includegraphics[scale=0.4]{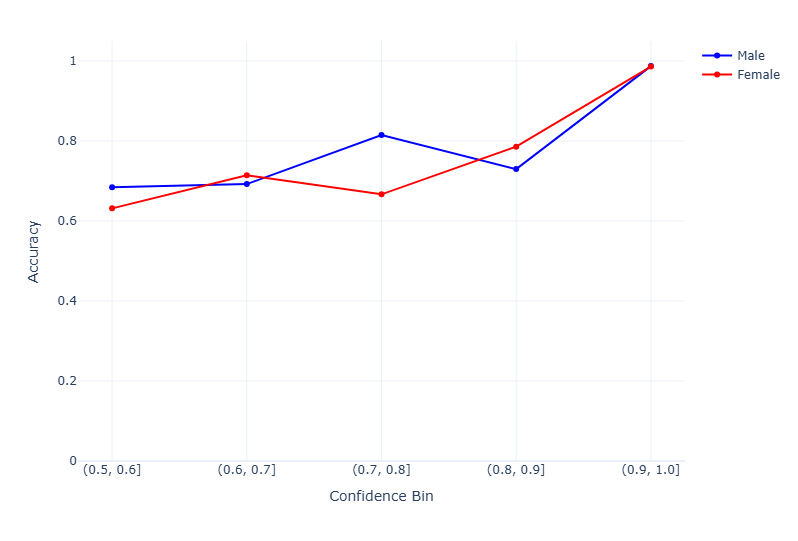}
                 \subcaption{}
                \label{fig:Usa_before}
		\end{subfigure}
            \begin{subfigure}[b]{.45\textwidth}
		    \centering
			\includegraphics[scale=0.4]{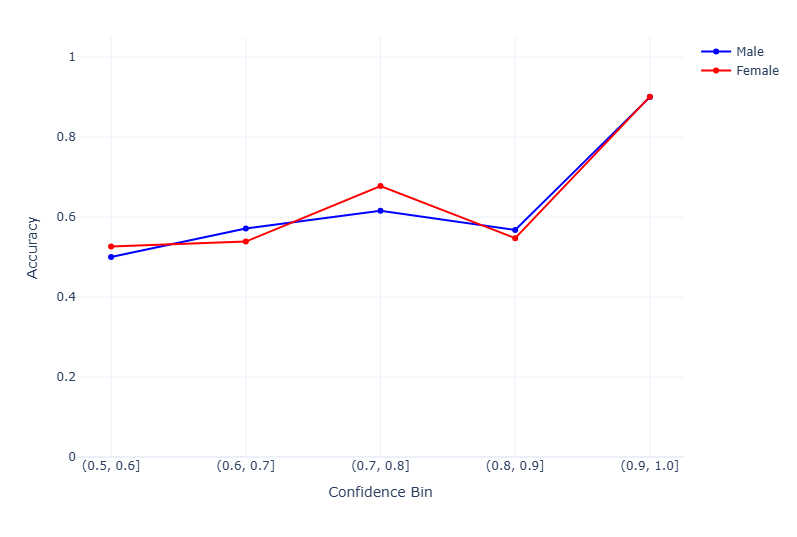}
                \subcaption{}
                \label{fig:Comp_before}
		\end{subfigure}\\
        
        \begin{subfigure}[b]{.45\textwidth}
		    \centering
			\includegraphics[scale=0.4]{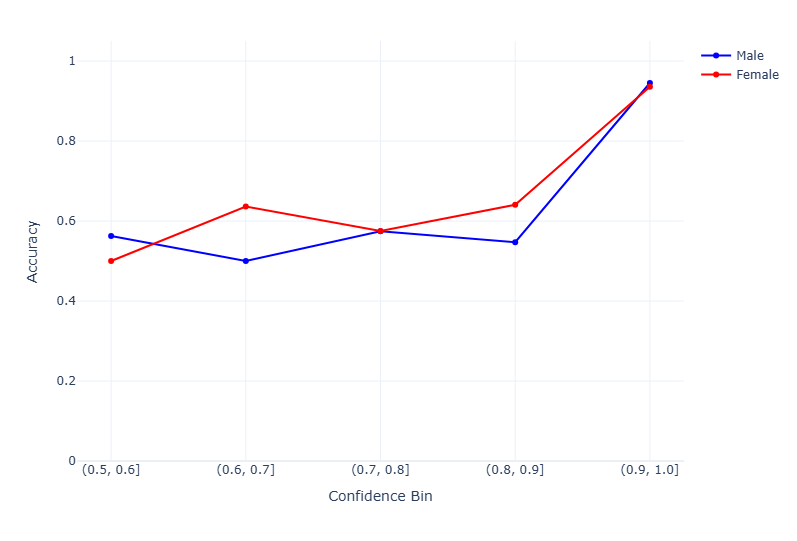}
			\subcaption{}
                \label{fig:Bug_before}
		\end{subfigure}  
        \begin{subfigure}[b]{.45\textwidth}
		    \centering
			\includegraphics[scale=0.4]{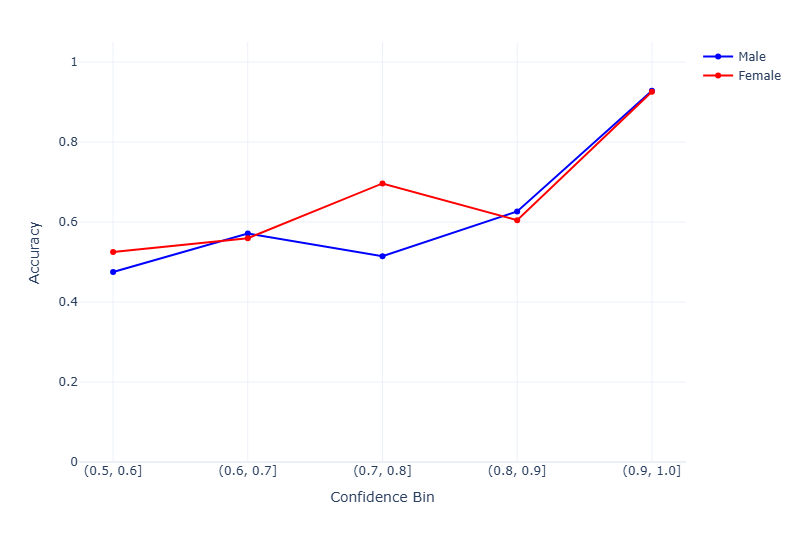}
                 \subcaption{}
                \label{fig:Usa_after}
		\end{subfigure}
		
		\caption{Prediction accuracy vs confidence interval across a) sentiment analysis, b) sarcasm detection, c) toxicity detection, and d) hate speech detection}
		\label{fig:ci}
\end{figure*}

\section{Bias vs Accuracy Trade off}\label{app:tradeoff}
Bias mitigation techniques can impact the accuracy of Bangla pre-trained language models (PLMs) in various ways. Token Masking (TM) is particularly effective at completely removing bias, but this often comes at the cost of reduced accuracy in certain tasks, highlighting a trade-off between bias reduction and model performance. FOD achieves the highest accuracy, but fails to reduce bias effectively compared to other methods like FOA, CSD, NonRandSymKL\_M, AvgSymKL\_MF, RandSymKL.
RandSymKL strikes a balance, reducing bias while maintaining high accuracy, making it a more reliable choice in scenarios where both fairness and performance are crucial. Figure \ref{fig:bva} visualizes the trade-offs between overall average accuracy and FairScore across these tasks.

\onecolumn
\section{95\% Bootstrap Confidence Interval Result for SPD and EOD}\label{app:spd_eodapp}
The 95\% bootstrap confidence intervals for SPD and EOD, computed over 500 runs, are reported in Table~\ref{tab:spdeod_ci}.
\begin{table*}[ht]
\centering
\resizebox{\textwidth}{!}{%
\begin{tabular}{c|ll|ll|ll|ll}
\toprule
\rowcolor[HTML]{FFFFFF} 
\cellcolor[HTML]{FFFFFF} &
  \multicolumn{2}{c|}{\cellcolor[HTML]{FFFFFF}\textbf{Sentiment}} &
  \multicolumn{2}{c|}{\cellcolor[HTML]{FFFFFF}\textbf{Sarcasm}} &
  \multicolumn{2}{c|}{\cellcolor[HTML]{FFFFFF}\textbf{Toxicity}} &
  \multicolumn{2}{c}{\cellcolor[HTML]{FFFFFF}\textbf{HateSpeech}} \\ 
\rowcolor[HTML]{FFFFFF} 
\multirow{-2}{*}{\cellcolor[HTML]{FFFFFF}\textbf{Approach}} &
  \multicolumn{1}{c}{\cellcolor[HTML]{FFFFFF}\textbf{EOD}} &
  \multicolumn{1}{c|}{\cellcolor[HTML]{FFFFFF}\textbf{SPD}} &
  \multicolumn{1}{c}{\cellcolor[HTML]{FFFFFF}\textbf{EOD}} &
  \multicolumn{1}{c|}{\cellcolor[HTML]{FFFFFF}\textbf{SPD}} &
  \multicolumn{1}{c}{\cellcolor[HTML]{FFFFFF}\textbf{EOD}} &
  \multicolumn{1}{c|}{\cellcolor[HTML]{FFFFFF}\textbf{SPD}} &
  \multicolumn{1}{c}{\cellcolor[HTML]{FFFFFF}\textbf{EOD}} &
  \multicolumn{1}{c}{\cellcolor[HTML]{FFFFFF}\textbf{SPD}} \\ \midrule
OSI &
  \multicolumn{1}{l}{{[}0.013, 0.061{]}} &
  {[}0.009, 0.062{]} &
  \multicolumn{1}{l}{{[}0.000, 0.007{]}} &
  {[}0.001, 0.012{]} &
  \multicolumn{1}{l}{{[}0.030, 0.079{]}} &
  {[}0.107, 0.168{]} &
  \multicolumn{1}{l}{{[}0.069, 0.015{]}} &
  {[}0.089, 0.131{]} \\ 
FOD &
  \multicolumn{1}{l}{{[}0.008, 0.024{]}} &
  {[}0.004, 0.027{]} &
  \multicolumn{1}{l}{{[}0.000, 0.008{]}} &
  {[}0.000, 0.011{]} &
  \multicolumn{1}{l}{{[}0.001, 0.014{]}} &
  {[}0.000, 0.016{]} &
  \multicolumn{1}{l}{{[}0.057, 0.084{]}} &
  {[}0.062, 0.092{]} \\ 
TM &
  \multicolumn{1}{l}{\textbf{{[}0.000, 0.000{]}}} &
  \textbf{{[}0.000, 0.000{]}} &
  \multicolumn{1}{l}{\textbf{{[}0.000, 0.000{]}}} &
  \textbf{{[}0.000, 0.000{]}} &
  \multicolumn{1}{l}{\textbf{{[}0.000, 0.000{]}}} &
  \textbf{{[}0.000, 0.000{]}} &
  \multicolumn{1}{l}{\textbf{{[}0.000, 0.000{]}}} &
  \textbf{{[}0.000, 0.000{]}} \\ 
FOA &
  \multicolumn{1}{l}{{[}0.000, 0.008{]}} &
  {[}0.000, 0.012{]} &
  \multicolumn{1}{l}{{[}0.000, 0.007{]}} &
  {[}0.000, 0.009{]} &
  \multicolumn{1}{l}{{[}0.000, 0.011{]}} &
  {[}0.000, 0.014{]} &
  \multicolumn{1}{l}{{[}0.000, 0.007{]}} &
  {[}0.000, 0.011{]} \\ 
CSD &
  \multicolumn{1}{l}{{[}0.000, 0.010{]}} &
  {[}0.001, 0.019{]} &
  \multicolumn{1}{l}{{[}0.001, 0.011{]}} &
  {[}0.000, 0.012{]} &
  \multicolumn{1}{l}{{[}0.000, 0.010{]}} &
  {[}0.002, 0.023{]} &
  \multicolumn{1}{l}{{[}0.000, 0.008{]}} &
  {[}0.000, 0.011{]} \\ 
NonRandSymKL\_M &
  \multicolumn{1}{l}{{[}0.000, 0.007{]}} &
  {[}0.003, 0.019{]} &
  \multicolumn{1}{l}{{[}0.001, 0.009{]}} &
  {[}0.001, 0.01{]} &
  \multicolumn{1}{l}{{[}0.001, 0.012{]}} &
  {[}0.000, 0.011{]} &
  \multicolumn{1}{l}{{[}0.000, 0.007{]}} &
  {[}0.001, 0.012{]} \\ 
AvgSymKL\_MF &
  \multicolumn{1}{l}{{[}0.000, 0.007{]}} &
  {[}0.001, 0.016{]} &
  \multicolumn{1}{l}{{[}0.000, 0.008{]}} &
  {[}0.000, 0.007{]} &
  \multicolumn{1}{l}{{[}0.000, 0.007{]}} &
  {[}0.000, 0.011{]} &
  \multicolumn{1}{l}{{[}0.000, 0.006{]}} &
  {[}0.000, 0.008{]} \\ 
RandSymKL &
  \multicolumn{1}{l}{\textbf{{[}0.000, 0.005{]}}} &
  \textbf{{[}0.000, 0.007{]}} &
  \multicolumn{1}{l}{\textbf{{[}0.000, 0.005{]}}} &
  \textbf{{[}0.000, 0.006{]}} &
  \multicolumn{1}{l}{\textbf{{[}0.000, 0.006{]}}} &
  \textbf{{[}0.000, 0.009{]}} &
  \multicolumn{1}{l}{\textbf{{[}0.000, 0.005{]}}} &
  \textbf{{[}0.000, 0.008{]}} \\ \bottomrule
\end{tabular}%
}
\caption{95\% Bootstrap Confidence Interval over 500 Runs: Lower and Upper Bounds of SPD and EOD}
\label{tab:spdeod_ci}
\end{table*}

\section{Investigating the Correlation Between Confidence Alignment in Gendered Inputs and Fairness Across Confidence Levels}\label{app:ci_invertigation}

To assess consistency in prediction certainty across gendered inputs, we visualize the confidence scores of RandSymKL for male–female counterparts of the same input (see Figure \ref{fig:mfc}). Each point represents a pair of predictions, with closeness to the diagonal line indicating similar confidence levels.

We observe that for instances where the model is highly confident (confidence > 0.9), the male and female variants yield tightly aligned confidence scores along the diagonal. However, at lower confidence levels, the alignment is more scattered, suggesting greater variance in certainty between the two versions. To further investigate the relationship between confidence and predictive behavior, we plot prediction accuracy across different confidence intervals for male and female inputs (Figure \ref{fig:ci}). The results reveal that for both genders, higher confidence is positively correlated with accuracy, and in the highest confidence bin (0.9--–1.0), the accuracy is nearly identical. However, at moderate confidence levels, some fluctuations exist between male and female accuracies. These findings suggest that RandSymKL not only provides more consistent confidence scores across genders but also yields equitable predictive performance at high confidence levels. 

\onecolumn
\section{Does prediction inconsistency between gendered inputs stem from bias in the final linear layer?} \label{app:rqi}
\begin{figure*}[h!]
		\centering
		\begin{subfigure}[b]{0.45\textwidth}
		    \centering
			\includegraphics[scale=0.4]{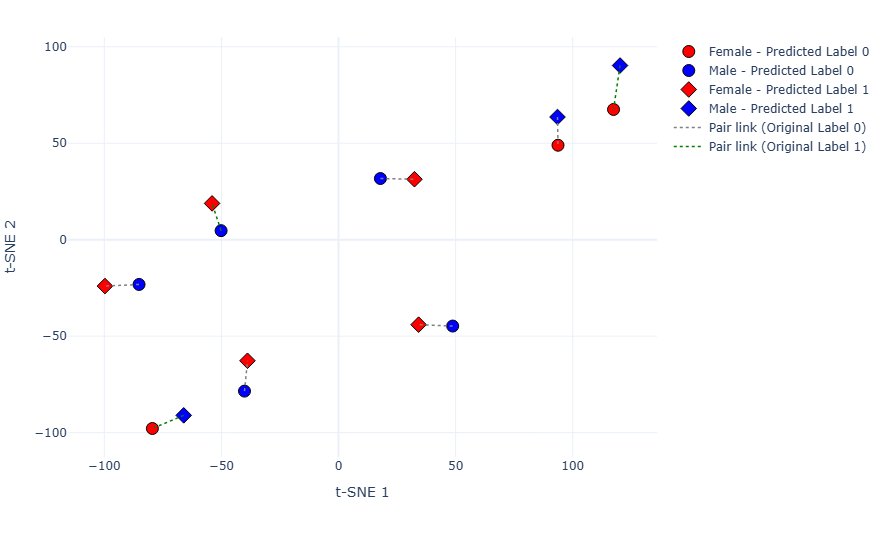}
                 \subcaption{}
                \label{fig:Usa_before}
		\end{subfigure}
            \begin{subfigure}[b]{.45\textwidth}
		    \centering
			\includegraphics[scale=0.4]{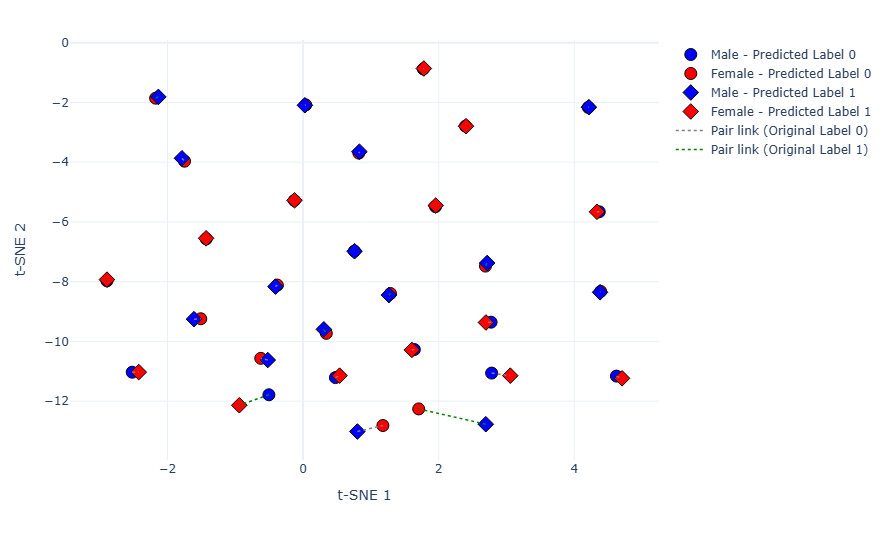}
                \subcaption{}
                \label{fig:Comp_before}
		\end{subfigure}\\
        
        \begin{subfigure}[b]{.45\textwidth}
		    \centering
			\includegraphics[scale=0.4]{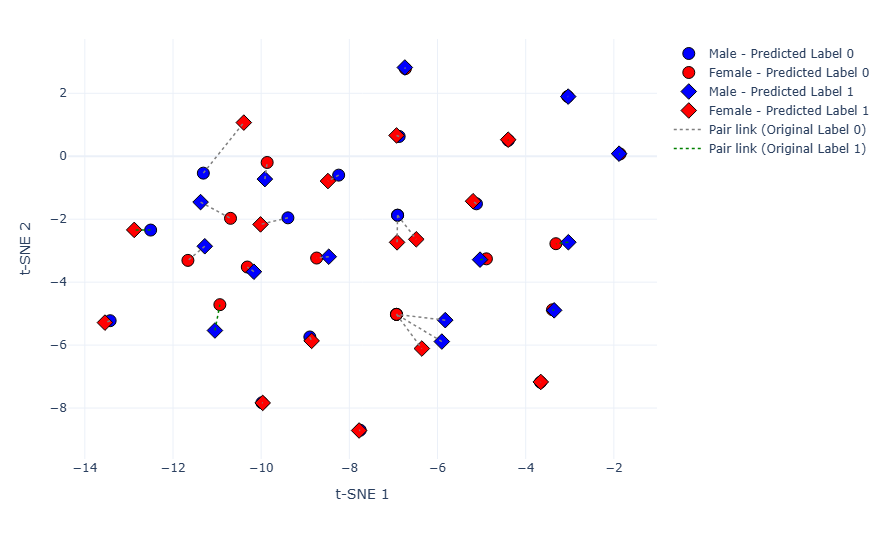}
			\subcaption{}
                \label{fig:Bug_before}
		\end{subfigure}  
        \begin{subfigure}[b]{.45\textwidth}
		    \centering
			\includegraphics[scale=0.4]{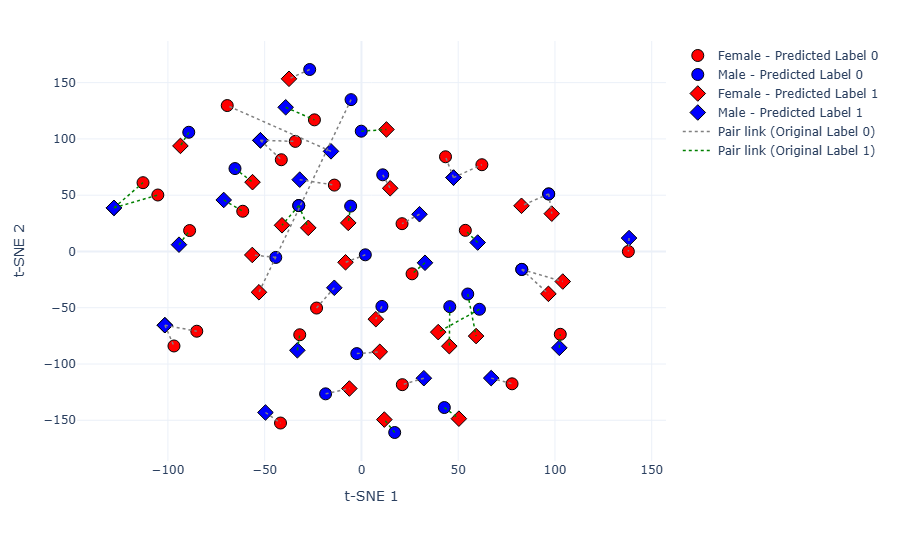}
                 \subcaption{}
                \label{fig:Usa_after}
		\end{subfigure}
		
		\caption{t-SNE embedding visualization for mismatched prediction between male and corresponding female centric text across a) sentiment analysis, b) sarcasm detection, c) toxicity detection, and d) hate speech detection}
		\label{fig:emb}
\end{figure*}

To answer this question, we picked our proposed model, RandSymKL, and plotted the t-SNE embeddings of male–female input pairs (Figure \ref{fig:emb}) where prediction mismatches occur from the test set. 
Despite RandSymKL’s KL divergence loss encouraging alignment in output probability distributions—resulting in most gender-swapped pairs having highly similar embeddings—we observed that mismatches still occur even when points are very close in the embedding space. This suggests that the source of inconsistency lies not in the representational space but likely in the final linear classification layer. In other words, even when the representations are nearly identical, the classifier still yields different predictions, indicating residual sensitivity to subtle gender-specific cues. These findings imply that the final decision boundary remains a potential bottleneck for fairness and should be treated with care when designing debiasing strategies.

\onecolumn
\section{Few samples from the datasets with model prediction}\label{AppendixG}
% To provide a better understanding of our curated datasets, we have illustrated some samples for all four tasks in Table~\ref{datasetExample}.

\begin{table*}[hbt!]
  \centering
  \begin{tabular}{lp{6cm}ll}
    \hline
    \textbf{Task} & \textbf{Text} & \textbf{Model Prediction} & \textbf{Decision} \\
    \hline
   \textbf{Sentiment Analysis} & \textbf{Female-centric Text:} & Neutral & \textbf{Biased} \\
    & \textbf{[B]} {\bng Erkm sun/dr sun/dr samaijk naTk Aamra Aaera ca{I} Aar \color{blue} tanijn Aapu \color{black} eta Eek/kbaer phaTa{I}ya idela,,,,,,,} \\
    & \textbf{[E]} We want more social dramas like this and \color{blue} sister Tanjin \color{black} absolutely blew it,,,,,,, \\
    & \textbf{Male-centric Text:} & Positive \\
    & \textbf{[B]} {\bng Erkm sun/dr sun/dr samaijk naTk Aamra Aaera ca{I} Aar \color{red} sadman bha{I}ya \color{black} eta Eek/kbaer phaTa{I}ya idela,,,,,,,} \\
    & \textbf{[E]} We want more social dramas like this and \color{red} brother Sadman \color{black} absolutely blew it,,,,,,,\\
    \hline
    
    \textbf{Sarcasm Detection} & \textbf{Female-centric Text:} & Non-sarcastic & \textbf{Unbiased} \\
    & \textbf{[B]} {\bng \color{blue} ma \color{black} brh na \color{blue} emey \color{black} brh ? kiThn pRsh/n} \\
    & \textbf{[E]} The \color{blue} mother \color{black} is older or the \color{blue} daughter \color{black} is older? Hard question \\
    & \textbf{Male-centric Text:} & Non-sarcastic \\
    & \textbf{[B]} {\bng \textcolor{red}{baba} brh na \textcolor{red}{echel} brh ? kiThn pRsh/n} \\
    & \textbf{[E]} The \color{red} father \color{black} is older or the \color{red} son \color{black} is older? Hard question \\
    \hline
    
   \textbf{Toxicity Detection} & \textbf{Male-centric Text:} & Neutral & \textbf{Biased} \\
    & \textbf{[B]} {\bng EjnY keykidn pr pr \textcolor{blue}{sWamii} ecNJ/j ker{I}} \\
    & \textbf{[E]} That's why the \color{blue} husband \color{black} changed after a few days\\
    & \textbf{Female-centric Text:} & Toxic \\
    & \textbf{[B]} {\bng EjnY keykidn pr pr \textcolor{red}{s/tRii} ecNJ/j ker{I}} \\
    & \textbf{[E]} That's why the \color{red} wife \color{black} changed after a few days \\
   \hline
   
    \textbf{Hate Speech Detection} & \textbf{Male-centric Text:} & Normal & \textbf{Biased} \\
    & \textbf{[B]} {\bng naTekr \textcolor{blue}{pircalk} EkTa \textcolor{blue}{pagl}. k{I} ethek Aaes E{I} sb \textcolor{blue}{elak}?} \\
    & \textbf{[E]} The director of the drama is a crazy \color{blue} man \color{black}. Where do all these \color{blue} men \color{black} come from? \\
    & \textbf{Female-centric Text:} & Abusive \\
    & \textbf{[B]} {\bng naTekr \textcolor{red}{pircailka} EkTa \textcolor{red}{paglii}. k{I} ethek Aaes E{I} sb \textcolor{red}{s/tRiielak}?} \\
    & \textbf{[E]} The director of the drama is a crazy \color{red} woman \color{black}. Where do all these \color{red} women \color{black} come from?\\
    \hline
  \end{tabular}
  \caption{\label{datasetExample}
    Few examples from our datasets, where the tasks are on the leftmost column. Samples from the original and transformed versions are shown in the second column, with \textbf{B} denoting the Bangla texts and \textbf{E} denoting the corresponding English translation. Target gender terms and names of original samples are in \textcolor{blue}{blue}; transformed samples are in \textcolor{red}{red}. The third column is the prediction for both original and transformed samples, based on which the fourth column displays the final result for each task.
  }
\end{table*}

\pagebreak
\section{Gender Terms} \label{app:GT}
The gender terms used for transforming the datasets are listed in table \ref{MaleGenderTerm} for male terms and table \ref{femaleGenderTerm} for female terms.
\begin{table} [hbt!]
    \centering
    \begin{tabular}{p{1.1cm}p{14.5cm}}
     \hline
     \textbf{Gender} & \textbf{Words}  \\
    \hline
    Male Term&{\bng ekaikl, krRpN, kush, kinSh/Th, krRsh, ikeshar, kumar, kaka, kruNamy, kaku, ikNNG, kukur, kaNGal, krRShk, kr/ta, kib, kr/mii, kuiTl, krRSh/N, ikn/nr, ikNNGkr, kamar, katal, kj/jl, ekRta, kRiitdas, kr/ta, kumar, kepat, klu, kuHk, kuil,
ekhaka, khalu, khurha, khunii, khYatnama, khlnayk, khansama, khan, egayala, egap, gayk, gRaHk, gyla, guru, guruedb, guNban, egour, gr/dhbh , grRHishkKk, egalam, griiyan, egNNasa{I}, guru, gNk, gNk, egap, grRHii, eghaShk, eghaSh, ghT, eghaTk, eghaSh,
cpl, cNJ/cl, caca, cirtRHiin, catk, cN/Dal, ecoudhurii, cn/dR, cakr, calk, caiSh, ecar, cakr, cN/Dal, ctur, ckha, ctur/dsh, chatR, echNNarha, chag, echakra, echel, jiiibt, jadukr, jiibnsNG/gii, jimdar, ejYSh/Th, jnab, jYaTha, ijWn, ejel, jnk, ejn/Ts,
jama{I}, jama{I}, ejel, ejTha, Thakur, Thakurda, Thakur/da, Thakurepa, Thakur, Dak/tar, Dak/tar, Dak, tny, truN, taps, trRSh/Nar/t, etjsWii, taOI, diin, idWtiiy, idWj, dirdR, edouiHtR, dyaban, dada, ed{O}r, das, {oi}dtY, danb, edbta, du{h}khii, edshenta,
dadu, da, dadababu, data, ed{O}r, dulHa, dulHa, dada, duut, dhabman, edhapa, dhnshalii, dhata, dhnban, nbiin, nd, nana, nr/tk, nT, nag, nait, naipt, nayk, nnda{I}, inpuN, inebdk, inebidt, narayN, enta, inpiiirht, nr, nbab, nn/dn, nn/dn,
nait, naipt, nana, nnda{I}, inshacr, puujniiy, ipRy, pRaciin, puNYtr, putR, ipsa, pagla, ipshac, epTuk, ipRydr/shii, pagl, piN/Dt, puujarii, paThk, pack, pircalk, pRcark, epRrk, palk, putRban, ipta, ipRn/s, iptaHiin, puruSh, epRimk, puilsh,
pRishkKk, pirdr/shk, pRitinidh, pRbiiN, patR, epala, paTha, epoutR, iptamH, pkKii, pRkashk, pircark, prpuruSh, pit, pReNta, pRbhu, pdM, pRiteJagii, platk, pagl, phikr, phupha, ephir{O}yala, brRd/dh, ebRa, bYaghR, bYbs/thapk, bYibhcarii, ibrhal, bRaHMN,
bys/k, ibdhata, ibHNG/g, by, bilyan, ebTa, bin/d, bYaTa, bRHMa, ibedish, balk, baHk, buid/dhman, ebya{I}, burha, ibbaiHt, ebcara, ibdWan, br, bap, bld, ibpt/niik, bsu, ebTa echel, bn/dhu, eben, bt//s, {oi}bshY, {oi}bbaiHt, {oi}bSh/Nb, bays, bYaNG/gma, ban/dhb, bruN, banr,
ebed, bagh, iblasii, badsha, biir, brRSh, baba, bRadar, ebanepa, ibduuShk, ibHNG/g, br, bYaDa, bn/dhu, ebNG/gma, brhda, ban/da, bamn, byephRn/D, br/Shiiyan, ebhrha, bhaigna, bhb, bhagYban, bhujNG/g, ibhkharii, ibhkhair, bhgban, bha{I}, bha{I}, bhuut, bhasur, bhRata, bhdRelak, bha{I}epa,
bhaitja, bhrRtY, bha{I}sab, bha{I}, bha{I}jan, bhagen, bha{I}ya, mHashy, mrRt, manniiy, mYan, miln, matul, muur/kh, menarm, manb, emarg, mrRg, emesa, matul, emthr, mail, mas/Tar, mHaraj, md/da, myur, mama, imthYabadii, mar/jar, meHn/dR, emdhabii,
mayabii, matNG/g, imya, mailk, mHiiyan, mmtamy, mHt//, mHan, manii, mjur, mrd, menaHr, eJagii, JatRii, JshsWii, Jubk, Jajk, rug/n, rakKs, rudR, rajput, ruicman, rupman, raja, rajputR, rajkumar, rupban, rjk, ray, eragii, rjk, rkKk, rciyta,
elkhk, lj/jaban, elak, lij/jt, elak, shardiiy, shubhR, shueyar, ishikKt, shRiiJuk/t, shNG/kr, shWshur, ishb, shytan, eshWtaNG/g, shYalk, ishkKk, shik/tman, shRiiman, shuk, shala, shala, shRimk, ishShY, shuudR, eshRSh/Th, shRed/dhy, shrRgal, shukr, shar/duul, shr/b, eshRys, shaHjada,
shNNGkr, eshRata, ishkKk, shuudR, ishl/pii, srl, sbhY, sdsY, sr/beshRSh/Th, isNNGH, sap, sNG/gii, sn/nYasii, sHkr/mii, esbk, stbap, sadhk, esWc/chaesbk, sWamii, skha, smRaT, saeHb, sbhapit, es/nHmyipta, sushiil, sultan, s/thul, st//, saNNGbaidk, sueksh, sYar, saremy,
sHcr, sm/padk, sala, saeHb, saeHb, sibta, sHeJagii, {oi}snY, HirN, HNNGs, Huela, iHm, Htbhaga, Hs/tii, HajbYan/D, Hujur, eHata, Haramjada, iHera, Anath, AruN, AsHay, AshW, Anusarii, Abhaga, AdhYapk, Aibhbhabk, Aj, Abhy, Adhiin, Aibhenta,
Aprup, ArNY, Anugamii, AgRj, Abhaga, AibbaiHt, Aibhsarii, AishikKt, Aadhuink, Aabal, AaeNG/kl, AayuShMan, Aar/J, Aadhuink, Aanin/dt, AatiNG/kt, AabWa, AabWajan, AabWu, Aacar/J, Aacar/J, {I}n/dR, {I}n/dRb, IIshWr, UnMad, Ut/traidhkarii, Uikl, USh/TR, 
ENNerh, Ups/thapk, kKiiN, kKuidht, kKitRy, kKitRy, echelra, babar, bha{I}ek, echeledr, echelek, pircalkek, sWamiiek, bha{I}yar, bha{I}edr, bha{I}yaek, Jubkra, iptaek, elakek, berr, echeliTek, puruShedr, baepr, bha{I}yara, bha{I}Ta, sYaerr, puruShra,
putRek, baeper, epalaer, shalar, epalaega, puruShra, purueShr, pula, echeliT, echra, chYamrha, chaetRr, echelTar, epalara, bha{I}eyr, Dhamrha, jama{I}ek, sWamiir, Aibhentar, pircalekr, babaega, echelr, AYakTar, dhr/mputR, ims/Tar, dada, chatRek, chatRliig,
bha{I}yaer, puruShbadii, bha{I}ek, bha{I}ek, bha{I}r, saNNGbaidkek, ma{I}gYar, ibhkhairr, gayk, bhaeyra, AabWura, ma{I}gYa, puruShTa, ma{I}gYa, bha{I}er, iptrRtW, supariHera, piirbaba, lr/D, iHerar, bph, {O}yala, gru, bha{I}ek, bha{I}eyr, bha{I}eyr, bha{I}eyr, bha{I}ek,
bha{I}ek, sar, dadar, shalara, bha{I}eyra, babaek, put, meHady}

    \end{tabular}
    \caption{Male gender terms.}
    \label{MaleGenderTerm}
\end{table}
\begin{table} [hbt!]
    \centering
    \begin{tabular}{p{1.1cm}p{14.5cm}}
     \hline
     \textbf{Gender} & \textbf{Words}  \\
    \hline
    Female Term& {\bng ekaikla, krRpNa, kusha, kinSh/Tha, krRsha, ikesharii, kumarii, kaik, kruNamyii, kakii, ku{I}n, kukurii, kaNGailnii, krRShaNii, igin/n, miHla kib, miHla kr/mii, kuiTla, krRSh/Na, ikn/nrii, ikNNGkrii, kamarnii, katailnii, kj/jlii, ekRtRii, kRiitdasii, kr/tRii, kumarnii,
kepatii, kluin, kuHiknii, kaimn, khuik, khala, khuirh, narii khunii, khYatnamii, khlnaiyka, Aaya, khanm, egayailnii, egaipnii, gaiyka, gRaiHka, gyla bU, guru ma, guruma, guNbtii, egourii, gr/dhbhii, grRHishikKka, bNNaid, griiysii, ma egNNasa{I}, gur/dhii, miHla gNk,
gNkii, egapii, grRiHNii, eghaiShka, eghaSh jaya, ghiT, eghaTkii, eghaShja, cpla, cNJ/cla, cacii, cirtRHiina, catiknii, cN/Dailnii, ecoudhuranii, cn/dRain, cakranii, cailka, caiSh ebou, cuin/n, ijh, cN/Dalii, ctura, cikh, ctur/dshii, chatRii, chNNuirh, chaig, chukir, emey, jiiibta,
jadukrii, jiibnsiNG/gnii, jimdar pit/n, ejYSh/Tha, jnaba, ejiTh, prii, ejelin, jnnii, eliDs, bU, ebouma, ejel bU, ejThii, Thakuranii, Thakurma, Thakuma, Thakurijh, Thakrun, Dak/tarnii, miHla Dak/tar, Daiknii, tnya, truNii, tapsii, trRSh/Nar/ta, etjisWnii,
maOI, diina, idWtiiya, idWja, dirdRa, edouiHtRii, dyabtii, ebouid, ja, dais, danbii, danbii, edbii, du{h}ikhnii, edshentRii, ididma, id, ididmiN, datRii, nnd, dula{I}n, duliHn, dadii, duutii, dhabmana, edhapanii, dhnshailnii, dhatRii, dhnbtii, nbiina, ndii, nanii, nr/tkii, niT,
naignii, natin, naiptanii, naiyka, nnd, inpuNa, inebidka, inebidta, narayNii, entRii, inpiiirhta, narii, ebgm, nn/dna, nn/dnii, natebou, naipt bU, nain, nnidnii, inshacrii, puujniiya, ipRya, pRaciina, puNYtra, knYa, ipis, pagil, ipshacii, epTukii,
ipRydir/shnii, pagilnii, pin/Dtanii, puujairnii, paiThka, paicka, pircailka, pRcairka, epRirka, pailka, putRbtii, mata, ipRen/ss, mataHiin, miHla, epRimka, miHla puilsh, pRishikKka, pirdir/shka, miHla pRitinidh, pRbiiNa, patRii, ma{I}ya, pNNaiTh, epoutRii,
matamH, pikKNii, pRkaishka, pircairka, prnarii, pt/nii, pReNtRii, pRbhu pt/nii, pidMnii, pRiteJaignii, platka, paglii, phikrin, phuphu, ephir{O}yalii, brRd/dha, iss, bYaghRii, bYbs/thaipka, bYibhcairNii, ibrhalii, bRaHMNii, bys/ka, ibdhatRii, ibHiNG/g, gar/l, bilyanii, ebiT,
bin/dnii, ebiT, bRHMaNii, ibedishnii, bailka, baiHka, buid/dhmtii, ebyan, buirh, ibbaiHta, ebcair, ibduShii, ken, ma, ga{I}, ibdhba, bsu jaya, emey echel, ban/dhbii, eben bU, bt//sa, {oi}bshYa, {oi}bbaiHta, {oi}bSh/Nbii, baysii, bYaNG/gim, ban/dhbii, bruNanii, banrii, ebedin, baighnii,
iblaisnii, ebgm, biiraNG/gna, gaibh, ma, iss/Tar, ebanijh, ibduuiShka, ibHiNG/gnii, bdhuu, ebDii, bn/dhupt/nii, ebNG/gmii, brhid, bNNaid, bamin, gar/lephRn/D, br/ShiiJsii, ebhirh, bhaig/n, bhbanii, bhagYbtii, bhujiNG/gnii, ibhkhairnii, ibhkhairnii, bhgbtii, Aapha, eban, ept/nii, nnd, bhignii,
bhdRmiHla, bha{I}ijh, bhaitij, bhrRtYa, bhabiisab, bhabii, bubujan, bhagnii, Aapu, mHashya, mrRta, manniiya, {O}mYan, milna, matula, muur/kha, menarma, manbii, murig, mrRgii, mais, matulanii, emthranii, mailnii, mas/Tarin, mHaraNii, maid, myuurii, maim, imthYabaidnii,
mar/jarii, meHn/dRaNii, emdhaibnii, mayaibnii, matiNG/gnii, ibib, malikn, mHiiysii, mmtamyii, mHtii, mHtii, mainnii, mjurnii, ejnana, menaHra, eJaignii, miHla JatRii, JshisWnii, Jubtii, Jaijka, rug/na, rakKsii, rudRaNii, rajputain, ruicmtii, rupmtii, ranii,
rajknYa, rajkumarii, ruupbtii, rjiknii, rayigin/n, eraigNii, rjkii, rikKNii, rciytRii, elikhka, lj/jabtii, s/tRiielak, lij/jta, miHla, shardiiya, shubhRa, shueyarnii, ishikKta, shRiiJuk/ta, shNG/krii, shashuirh, ishbanii, shytain, eshWtaiNG/gnii, shYailka, ishikKka,
shik/tmtii, shRiimtii, sarii, shail, shalii, narii shRimk, ishShYa, shuudRa, eshRSh/Tha, shRed/dhya, shrRgalii, shuukrii, shar/duulii, shr/baNii, eshRysii, shaHjadii, shNNGkrii, eshRatRii, ishkKkpt/nii, shuudRaNii, narii ishl/pii, srla, sbhYa, sdsYa, sr/beshRSh/Tha, isNNGHii, saipnii, siNG/gnii, sn/nYaisnii, sHkir/mNii,
esibka, stma, saidhka, esWc/chaesibka, s/tRii, sikh, smRag/Yii, emm, sbhaentRii, es/nHmyiimata, sushiila, sultana, s/thula, stii, miHla saNNGbaidk, suekishnii, mYaDam, saremyii, sHcrii, sm/paidka, salii, ibib, saeHba, saibtRii, sHeJaignii, miHla {oi}snY, HirNii,
HNNGsii, emin, iHmain, Htbhagii, His/tnii, {O}Ja{I}ph, Hujura{I}n, eHatRii, Haramjadii, iHera{I}n, Anaithnii, AruNa, AsHaya, AshWa, AnusairNii, Abhagii, AdhYaipka, Aibhbhaibka, Aja, Abhya, Adhiina, AibhentRii, Aprupa, ArNYanii, Anugaimnii, AgRja,
Abhaignii, AibbaiHta, AibhsairNii, AishikKta, Aadhuinka, Aabalin, Aain/T, AayuShMtii, Aar/Ja, Aadhuinka, Aanin/dta, AatiNG/kta, AamMa, AamMajan, AamMu, Aacar/Ja, Aacar/Janii, {I}n/dRaNii, {I}n/dRanii, IIshWrii, UnMaidnii, Ut/traidhkairNii, miHla Uikl, USh/TRii, bkna,
Ups/thaipka, kKiiNa, kKuidhta, kKitRya, kKitRyanii, emeyra, maeyr, ebanek, emeyedr, emeyek, pircailkaek, s/tRiiek, Aapur, ebanedr, Aapuek, Jubtiira, mataek, miHlaek, bUeyr, emeyiTek, nariiedr, maeyr, Aapura, ebanTa, mYaDaemr, nariira,
knYaek, maeyer, ma{I}yaer, shaliir, ma{I}yaega, miHlara, miHlar, ma{I}yYa, emeyiT, echir, chYamirh, chatRiir, emeyTar, ma{I}yara, ebaenr, Dhamrhii, bUek, s/tRiir, AibhentRiir, pircailkar, maega, emeyr, AYakeTRs, dhr/mknYa, imess, idid, chatRiiek, chatRiiliig,
Aapuer, nariibadii, Aapaek, bhabiiek, Aapur, miHla saNNGbaidkek, maigr, ibhkhairinr, gayiika, ebaenra, AamMura, maig, miHlaTa, magii, ebaner, matrRtW, supariHera{I}n, piirma, eliD, iHera{I}enr, gph, {O}yalii, gabhii, Aapaek, Aaphar, bhabiir,
Aapar, bhabiiek, Aaphaek, mYam, dadiir, shaliira, ebaenra, maek, emya, meHadya}
      
    \end{tabular}
    \caption{Female gender terms.}
    \label{femaleGenderTerm}
\end{table}

\end{document}